\documentclass{ieeeaccess}
\usepackage{cite}
\usepackage{amsmath,amssymb,amsfonts}
\usepackage{algorithmic}
\usepackage{graphicx}
\usepackage{textcomp}
\usepackage{multirow}
\usepackage{caption}
\def\BibTeX{{\rm B\kern-.05em{\sc i\kern-.025em b}\kern-.08em
    T\kern-.1667em\lower.7ex\hbox{E}\kern-.125emX}}

\usepackage{soul}
\soulregister\cite7
\soulregister\ref7
\begin{document}
\history{Date of publication xxxx 00, 0000, date of current version xxxx 00, 0000.}
\doi{10.1109/ACCESS.2017.DOI}

\title{Diversity in Machine Learning}
\author{\uppercase{Zhiqiang Gong}\authorrefmark{1},
\uppercase{Ping Zhong\authorrefmark{1}, \IEEEmembership{Senior Member, IEEE}, and Weidong Hu\authorrefmark{1}}}
\address[1]{National Key Laboratory of Science and Technology on ATR,  College of Electronic Science and Technology, National University of Defense Technology, Changsha 410073, China (e-mail: gongzhiqiang13@nudt.edu.cn, zhongping@nudt.edu.cn, wdhuatr@icloud.com)}
\tfootnote{This work was supported in part by the Natural Science Foundation of China under Grant 61671456 and 61271439, in part by the Foundation for the Author of National Excellent Doctoral Dissertation of China (FANEDD) under Grant 201243, and in part by the Program for New Century Excellent Talents in University under Grant NECT-13-0164.}

\markboth
{Gong \headeretal: Diversity in Machine Learning}
{Gong \headeretal: Diversity in Machine Learning}

\corresp{Corresponding author: Ping Zhong (e-mail: zhongping@nudt.edu.cn).}

\begin{abstract}
Machine learning methods have achieved good performance and been widely applied in various real-world applications. They can learn the model adaptively and be better fit for special requirements of different tasks. Generally, a good machine learning system is composed of plentiful training data, a good model training process, and an accurate inference.
Many factors can affect the performance of the machine learning process, among which the diversity of the machine learning process is an important one.  The diversity can help each procedure to guarantee a total good machine learning: diversity of the training data ensures that the training data can provide more discriminative information for the model, diversity of the learned model (diversity in parameters of each model or diversity among different base models) makes each parameter/model capture unique or complement information  and the diversity in inference can provide multiple choices each of which corresponds to a specific plausible local optimal result. Even though the diversity plays an important role in machine learning process, there is no systematical analysis of the diversification in machine learning system. In this paper, we systematically summarize the methods to make data diversification, model diversification, and inference diversification in the machine learning process, respectively.  In addition, the typical applications where the diversity technology improved the machine learning performance have been surveyed, including the remote sensing imaging tasks, machine translation, camera relocalization, image segmentation, object detection, topic modeling, and others. Finally, we discuss some challenges of the diversity technology in machine learning and point out some directions in future work. Our analysis provides a deeper understanding of the diversity technology in machine learning tasks, and hence can help design and learn more effective models {for real-world applications}.
\end{abstract}

\begin{keywords}
Diversity, Training Data, Model Learning, Inference, Supervised Learning, Active Learning, Unsupervised Learning, Posterior Regularization
\end{keywords}

\titlepgskip=-15pt

\maketitle

\section{Introduction}\label{intro}

Traditionally, machine learning methods can learn {model's} parameters automatically with {the} training samples and thus it can {provide models with good performances which can satisfy the special requirements of various applications.} Actually, it has achieved great success in tackling many real-world artificial intelligence and data mining problems \cite{128}, such as object detection \cite{54, 55}, natural image processing \cite{32}, autonomous car driving \cite{130}, urban scene understanding \cite{urban}, machine translation \cite{103}, and web search/information retrieval \cite{155}, {and others}. A success machine learning system often requires plentiful training data which can provide enough information to train the model, a good model learning process which can better model the data, and an accurate inference to discriminate different objects.
However, in real-world applications, limited number of labelled training data are available. Besides, {there exist large amounts of parameters in the machine learning model. These} would make the "over-fitting" phenomenon  in the machine learning process. Therefore, {obtaining an accurate inference  from the machine learning model tends to be a difficult task.}
Many factors can help to improve the performance of the machine learning process, among which the diversity in machine learning plays an important role.

{Diversity shows different concepts depending on context and application \cite{34}.
Generally, a diversified system contains more information and can better fit for various environments. It has already become an important property in many social fields, such as biological system, culture, products and so on. Particularly, the diversity property also has significant effects on the learning process of the machine learning system.}
{Therefore, we wrote this survey mainly for two reasons. First, while the topic of diversity in machine learning methods has received attention for many years, there is no framework of diversity technology on general machine learning models. Although Kulesza et al. discussed the determinantal point processes (DPP) in machine learning which is only one of the measurements for diversity\cite{34}.
\cite{add_40} mainly summarized the diversity-promoting methods for obtaining multiple diversified search results in the inference phase.
Besides, \cite{add_3,add_14,30} analyzed several methods on classifier ensembles, which represents only a specific form of ensemble learning. All these works do not provide a full survey of the topic, nor do they focus on machine learning with general forms. Our main aim is to provide such a survey, hoping to induce diversity in general machine learning process. As a second motivation, this survey is also useful to researchers working on designing effective learning process.}

{Here, the diversity in machine learning works mainly on decreasing the redundancy between the data or the model and providing informative data or representative model in the machine learning process.}
{This work will discuss the diversity property from} different components of the machine learning process, including the training data, {the learned model}, {and} the inference.
{ The diversity in machine learning tries to decrease the redundancy in the training data, the learned model as well as the inference and provide more information for machine learning process. It can improve the performance of the model and has  played an important role in machine learning process. In this work, we} summarize the diversification of machine learning into three categories: the diversity in training data (data diversification), the diversity of the model/models (model diversification) and the diversity of the  inference (inference diversification).

\textit{Data diversification} can provide samples with enough information to train the machine learning model. The diversity in training data { aims to }maximize the information contained in the data. Therefore, the model can learn more information {from the data via the learning process} and the learned model can {be better fit for the data}. Many prior works have imposed the diversity on the construction of each training batch for the machine learning process to train the model more effectively \cite{36}. In addition, diversity in active learning can also make the labelled training data contain the most information \cite{1,3} and thus the learned model can achieve good performance with limited training samples. Moreover, in special unsupervised learning method by \cite{gong_ijcnn}, diversity of the pseudo classes can encourage the classes to repulse from each other and thus the learned model can provide more discriminative features from the objects.

\textit{Model diversification} comes from the diversity in human visual system. \cite{66, 67, 68} have shown that the human visual system represents decorrelation and sparseness, namely diversity. This makes different neurons in the human learning respond to different stimuli and generates little redundancy in the learning process which ensures the high effectiveness of the human learning. However, general machine learning methods usually perform the redundancy in the learned model where different factors model the similar features \cite{44}. Therefore, {\textit{diversity between the parameters of the model (D-model)}} could significantly improve the performance of the machine learning systems.
The D-model tries to encourage different parameters in each model to be diversified and each parameter can model unique information \cite{14, 26}. As a result, the performance of each model can be significantly improved \cite{13}.
However, general machine learning model usually provides a local optimal representation of the data with limited training data. Therefore, ensemble learning, which can learn multiple models simultaneously, becomes another hot machine learning methods to provide multiple choices and {has been widely applied in many real-world applications, such as the speech recognition \cite{131, 132}, and image segmentation \cite{99}}. However, general ensemble learning usually makes the learned multiple base models converge to the same or similar local optima. Thus,
\textit{diversity among multiple base models by ensemble learning (D-models) }, which tries to repulse different base models and encourages each base model to provide choice reflecting multi-modal belief \cite{22,99,114}, can provide multiple diversified choices and significantly improve the performance.

Instead of learning multiple models with D-models, one can also obtain multiple choices in the inference phase, {which is generally called multiple choice learning (MCL)}. However, the obtained choices from usual machine learning systems presents similarity between each other where the next choice will be one-pixel shifted versions of others \cite{23}. Therefore, to overcome this problem, {diversity-promoting prior can be imposed over the obtained multiple choices from the inference}. Under the
\textit{inference diversification}, the model  can provide choices/representations with more complement information \cite{5,9,12,20}. This could further improve the performance of the machine learning process and provide { multiple discriminative choices} of the objects.

This work systematically covers the literature on diversity-promoting methods over data diversification, model diversification, and inference diversification in machine learning tasks. In particular, three main questions from the analysis of diversity technology in machine learning have arisen.
\begin{itemize}
\item	How to measure the diversity of  the training data, the learned model/models, and the inference and enhance these diversity in machine learning system, respectively?
How do these methods work on the diversification of the machine learning system?

\item	Is there any difference between the diversification of the model and models? Furthermore, is there any similarity between the diversity in the training data, the learned model/models, and the inference? 
\item  Which real-world applications can the diversity be applied in to improve the performance of the machine learning models? How do the diversification methods work on these applications?
\end{itemize}

Although all of the three problems are important, none of them has been thoroughly answered. Diversity in machine learning can balance the training data, encourage the learned parameters to be diversified, and {diversify the multiple choices from the inference.} Through enforcing diversity in the machine learning system, the machine learning model can present a better performance.  Following the framework, the three questions above have been answered with both the theoretical analysis and the real-world applications.

\begin{figure}
\centering
  \includegraphics[width=0.49\textwidth]{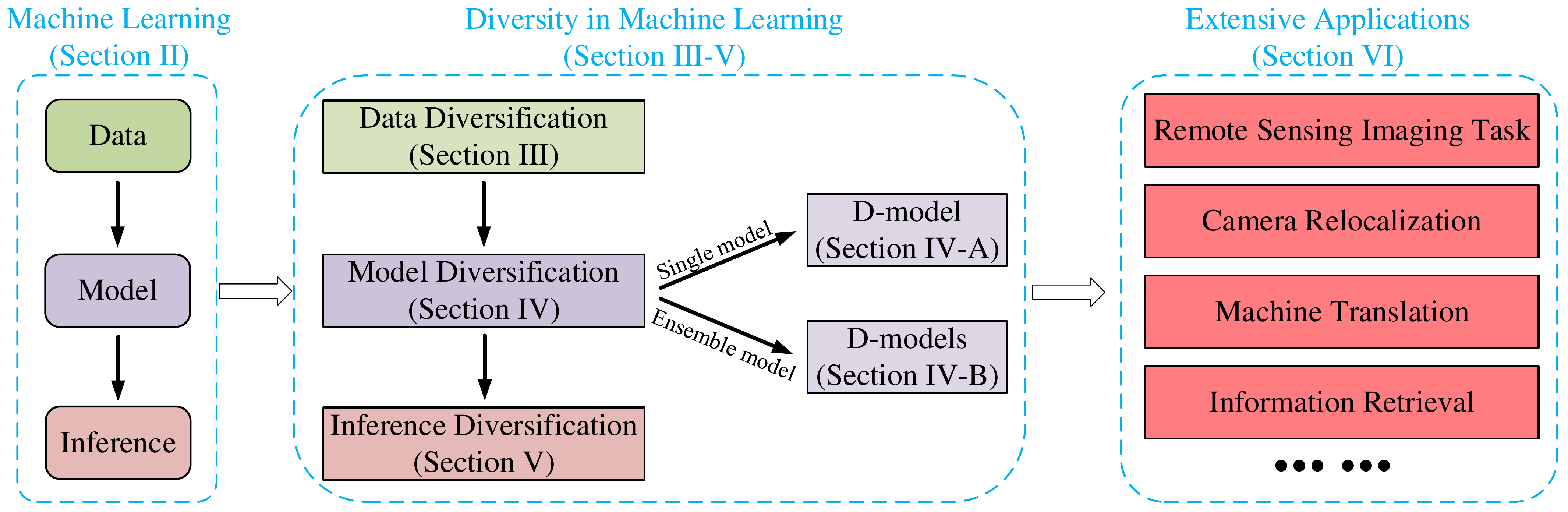}
\caption{{The basic framework of this paper. The main body of this paper consists of three parts: General Machine Learning Models in Section II, Diversity in Machine Learning in Section III-V, and Extensive Applications in Section VI.}}
\label{fig:article_structure}       
\end{figure}

{The remainder of this paper is organized as Fig. \ref{fig:article_structure} shows.} Section \ref{sec:general} discusses the general forms of the supervised learning and the active learning as well as a special form of unsupervised learning in machine learning model. {Besides,
as Fig. \ref{fig:article_structure} shows, Sections \ref{sec:data}, \ref{sec:model} and \ref{sec:inference} introduce the diversity methods in machine learning models.}
Section \ref{sec:data} outlines some of the prior works on diversification in training data. Section \ref{sec:model} reviews the strategies for model diversification, including the D-model, and the D-models. The prior works for inference diversification are summarized in Section \ref{sec:inference}. {Finally, section \ref{sec:application} introduces some applications of the diversity-promoting methods in prior works, and then we do some discussions, conclude the paper and point out some future directions.}

\begin{figure*}
\centering
  \includegraphics[width=0.85\textwidth]{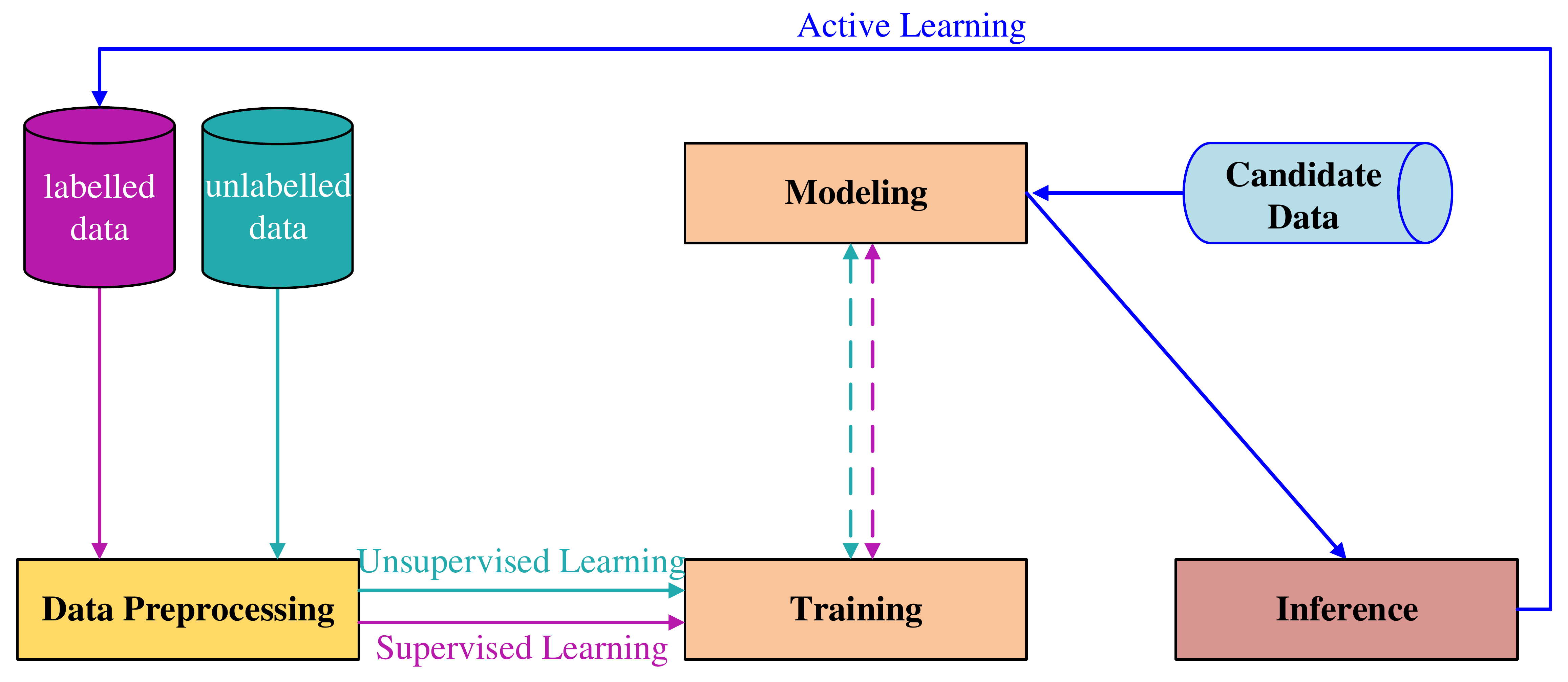}
\caption{Flowchart for training process of general machine learning (including the active learning, supervised learning and unsupervised learning). We can find that when the training data is labelled, the training process is supervised. In contrast, the training process is unsupervised. Besides, it should be noted that when both the labelled and unlabelled data are used for training, the training process is semi-supervised.}
\label{fig:01}       
\end{figure*}

\section{General Machine Learning Models}\label{sec:general}
Traditionally, machine learning consists of supervised learning, active learning, unsupervised learning, and reinforcement learning. For reinforcement learning, training data is given only as the feedback to the program's actions in a dynamic environment, and it does not require accurate input/output pairs and the sub-optimal actions need not to be explicitly correct. However, the diversity technologies mainly work on the model itself to improve the model's performance. Therefore, this work will ignore the reinforcement learning and mainly discuss the machine learning model as Fig. \ref{fig:01} shows.  In the following, we'll introduce the general forms of supervised learning and {a representative form of active learning as well as a special form of unsupervised learning.}

\subsection{Supervised Learning}\label{subsec:machine}
We consider the task of general supervised machine learning models, which are commonly used in real-word machine learning tasks. Fig. \ref{fig:01} shows the flowchart of general machine learning methods in this work. As Fig. \ref{fig:01} shows, the supervised machine learning model consists of data pre-processing, training (modeling), and inference. All of the steps can affect the performance of the machine learning process.

Let $X=\{{\bf x}_1,{\bf x}_2, \cdots ,{\bf x}_{N_1}\}$ denote the set of training samples and $y_i$ is the corresponding label of ${\bf x}_i$, where {$y_i \in \Omega=\{cl_1, cl_2, \cdots, cl_n\}$ ($\Omega$ is the set of class labels, $n$ is the number of the classes, and $N_1$ is the number of the labelled training samples)}.
{Traditionally, the machine learning task can be formulated as the following optimization problem \cite{157,158}:}
\begin{equation}\label{eq:01}
\begin{aligned}
&\max_W L(W|X)\\
&s.t.\ g(W)\geq 0
\end{aligned}
\end{equation}
where {$L(W|X)$ represents the loss function and $W$ is the parameters of the machine learning model. Besides, $g(W)\geq 0$ is the constraint of the parameters of the model.
Then, }the Lagrange multiplier of the optimization can be reformulated as follows.
\begin{equation}\label{eq:02}
L_0=L(W|X)+\eta g(W)
\end{equation}
where $\eta$ is a positive value. Therefore, the machine learning problem can be seen as the minimization of $L_0$.

\begin{figure*}
\centering
  \includegraphics[width=0.9\textwidth]{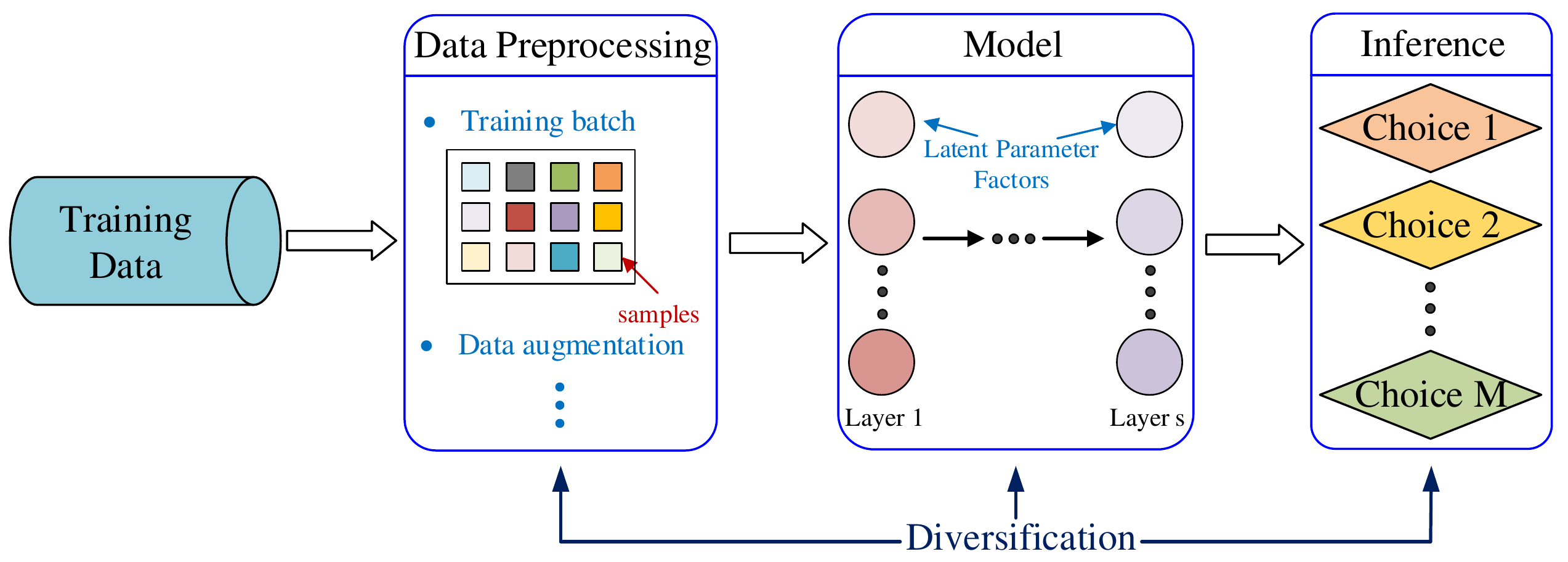}
\caption{{Flowchart of a special form of  supervised machine learning with single model. Since diversity mainly occurs in the training batch in the data-preprocessing, this work will mainly discuss the diversity of samples in the training batch for data diversification. Generally,} the more diversification and balance  each training batch is, the more effectiveness the training process is. In addition, it should be noted that the factors in the same layer of the model can be diversified to improve the representational ability of the model (which is  called D-model in this paper). Moreover, when we obtain multiple choices from the model, the obtained choices are desired to provide more complement information. Therefore, some works focus on the diversification of multiple choices (which we call inference diversification).}
\label{fig:02}       
\end{figure*}

\begin{figure*}
\centering
  \includegraphics[width=0.9\textwidth]{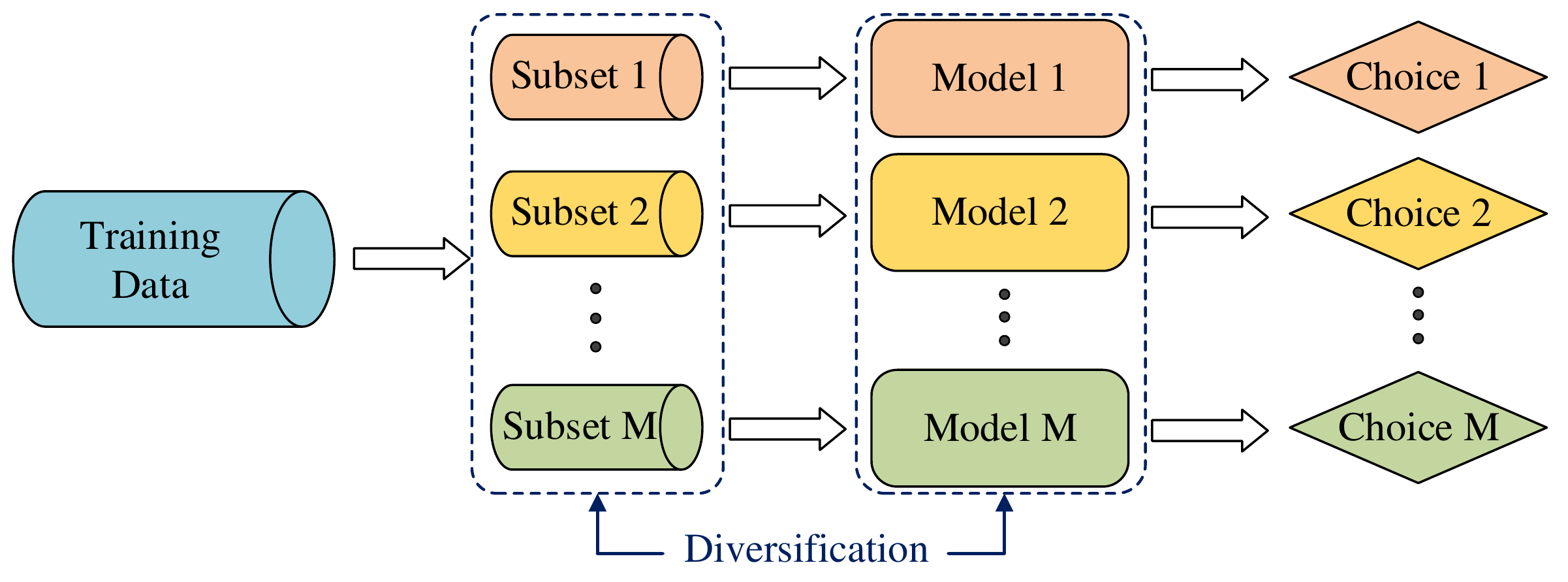}
\caption{Flowchart of supervised machine learning with multiple parallel models. We can find that a best strategy to diversify the training set for different models can improve the performance of multiple models. Furthermore, we can diversify different models directly to enforce different model to provide more complement information for further analysis.}
\label{fig:03}       
\end{figure*}

Figs. \ref{fig:02} and \ref{fig:03} show {the flowchart of two special forms} of supervised learning models, which are generally used in real-world applications. Among them, Fig. \ref{fig:02} shows the {flowchart of a special form of} supervised machine learning with a single model. Generally, in the data-preprocessing stage, the more diversification and balance each training batch has, the more effectiveness the training process is. In addition, it should be noted that the factors in the same layer of the model can be diversified to improve the representational ability of the model (which is called D-model in this paper). Moreover, when we obtain multiple choices from the model in the inference, the obtained choices are desired to provide more complement information. Therefore, some works focus on the diversification of multiple choices (which we call inference diversification). Fig. \ref{fig:03} shows the flowchart of supervised machine learning with multiple parallel base models. We can find that a best strategy to diversify the training set for different base models can improve the performance of the whole ensemble (which is called D-models). Furthermore, we can diversify these base models directly to enforce each base model to provide more complement information for further analysis.

\subsection{Active Learning}\label{sec:active}

Since labelling is always cost and time consuming, it usually cannot provide enough labelled samples for training in real world applications. Therefore, active learning, which can reduce the label cost and keep the training set in a moderate size, plays an important role in the machine learning model \cite{2}. {It can make use of the most informative samples and  provide a higher performance with less labelled training samples.}

Through active learning, we can choose the most informative samples for labelling to train the model. This paper will take the Convex Transductive Experimental Design (CTED) as a representative of the active learning methods \cite{69, 70}.

Denote {$U=\{{\bf u}_i\}_{i=1}^{N_2}$} as the candidate unlabelled samples for active learning, {where $N_2$ represents the number of the candidate unlabelled samples.}
Then, the active learning problem can be formulated as the following optimization problem \cite{70}:
\begin{equation}\label{03}
\begin{aligned}
& A^*, {\bf b}^* = \arg\min\limits_{A,{\bf b}} \|U-UA\|_F^2+\sum_{i=1}^{N_2} \frac{\sum_{j=1}^{N_2} a_{ij}^2}{b_i}+\alpha \|{\bf b}\|_1 \\
&s.t.\ b_i \geq 0, i=1,2,\cdots, N_2
\end{aligned}
\end{equation}
where $a_{ij}$ is the $(i, j)$-th entry of $A$, and $\alpha$ is a positive tradeoff parameter. {$\|\cdot\|_F$ represents the Frobenius norm (F-norm) which calculates the root of the quadratic sum of the items in a matrix. } As is shown, CTED utilizes {a data reconstruction framework to select the most informative samples for labelling}. The matrix $A$ contains reconstruction coefficients and ${\bf b}$ is the sample selection vector. The $L_1$-norm makes the learned ${\bf b}$ to be sparse.
Then, the obtained ${\bf b}^*$ is used to select samples for labelling and finally the training set is constructed with {the selected training samples.}
However, the selected samples from CTED usually make similarity from each other, which leads to the redundancy of the training samples. Therefore, diversity property is also required in the active learning process.

\subsection{Unsupervised Learning}\label{subsec:unsupervised}

As discussed in former subsection, limited number of the training samples will limit the performance of the machine learning process. Instead of the active learning, to solve the problem, {unsupervised learning methods provide another way to train the machine learning model without the labelled training samples.} This work will mainly {discuss a special unsupervised learning process developed by \cite{gong_ijcnn}}, which is an end-to-end self-supervised method.

Denote ${\bf c}_i(i=1,2,\cdots, \Lambda)$ as the center points which is used to formulate the pseudo classes in the training process where $\Lambda$ represents the number of the pseudo classes. {Just as subsection \ref{sec:active},  $U=\{{\bf u}_1,{\bf u}_2, \cdots ,{\bf u}_{N_2}\}$ represents the unlabelled training samples} and {$N_2$ denotes the number of the unsupervised samples}. {Besides, denote $\varphi({\bf u}_i)$ as the features of ${\bf u}_i$ extracted from} the machine learning model. Then, the pseudo label $z_i$ of the data { ${\bf u}_i$} can be defined as
\begin{equation}\label{eq:un_1}
  z_i=\min\limits_{k\in \{1,2,\cdots,\Lambda\}} \|{\bf c}_k-\varphi({{\bf u}_i})\|,
\end{equation}
Then, the problem can be transformed to a supervised one with the pseudo classes.
As shown in subsection \ref{subsec:machine}, the machine learning task can be formulated as the following optimization \cite{gong_ijcnn}
\begin{equation}\label{eq:un_2}
  \max\limits_{W,{\bf c}_i} L(W|U,z_i) + \eta g(W) + \sum_{k=1}^{N_2}\|{\bf c}_{z_k}-\varphi({\bf u}_k)\|
\end{equation}
{where $L(W|U,z_i (i=1,2,\cdots,N_2))$ denotes the optimization term and  $\sum_{k=1}^{N_2}\|{\bf c}_{z_k}-\varphi({\bf u}_k)\|$} is used to minimize the intra-class variance of the constructed pseudo-classes. $g(W)$ demonstrates the constraints in Eq. \ref{eq:02}.
With the iteratively learning of Eq. \ref{eq:un_1} and Eq. \ref{eq:un_2}, the machine learning model can be {trained unsupervisedly}.

{Since the center points play an important role in the construction of the pseudo classes, diversifying these center points and repulsing the points from each other can better discriminate these pseudo classes.} This would show positive effects on improving the effectiveness of the unsupervised learning process.

\subsection{Analysis}\label{sec:analysis}
As former subsections show, diversity can improve the performance of the machine learning process. In the following,
this work will summarize the diversification in machine learning from three aspects: data diversification, model diversification, and inference diversification.

To be concluded, diversification can be used in supervised learning, active learning, and unsupervised learning to improve the model's performance. According to the models in \ref{subsec:machine} and \ref{sec:active}, the diversification technology in machine learning model has been divided into three parts: data diversification (Section \ref{sec:data}), model diversification (Section \ref{sec:model}), and inference diversification (Section \ref{sec:inference}). Since the diversification in training batch (Fig. \ref{fig:02}) and the diversification in active learning and unsupervised learning mainly consider the diversification in training data, we  summarize the prior works in these diversification as data diversification in section \ref{sec:data}. Besides, the diversification of the model in Fig. \ref{fig:02} and {the multiple base models in Fig. \ref{fig:03} mainly focus on the diversification in the machine learning model directly,} and thus we summarize these works as model diversification in section \ref{sec:model}. Finally, the inference diversification in Fig. \ref{fig:02} will be summarized in section \ref{sec:inference}. In the following section, we'll first introduce the data diversification in machine learning models.

\section{Data Diversification}\label{sec:data}

{Obviously, the training data plays an important role in the training process of the machine learning models. For supervised learning in subsection \ref{subsec:machine}, the training data provides more plentiful information for the learning of the parameters. While for active learning in subsection \ref{sec:active}, the learning process would select the most informative and less redundant samples for labelling to obtain a better performance. Besides, for unsupervised learning in subsection \ref{subsec:unsupervised}, the pseudo classes can be encouraged to repulse from each other and the model can provide more discriminative features unsupervisedly. The following will introduce the methods for these data diversification in detail.}

\subsection{Diversification in Supervised Learning} \label{subsec:dpp}

General supervised learning model is usually trained with mini-batches to accurately estimate the training model. Most of the former works generate the mini-batches randomly. {However, due to the imbalance of the training samples under random selection, redundancy may occur in the generated mini-batches which shows negative effects on the effectiveness of the machine learning process.} Different from classical stochastic gradient descent (SGD) method which relies on the uniformly sampling data points to form a mini-batch, {\cite{36, add_12}} proposes a non-uniformly sampling scheme based on the {determinantal point process (DPP)} measurement.

A DPP is a distribution over subsets of a fixed ground set, which prefers {a diverse set of data} other than a redundant one \cite{34}.
{Let $\Theta$ denote a continuous space and the data ${\bf x}_i\in \Theta (i=1,2,\cdots,N_1)$.} Then, the DPP denotes a positive semi-definite kernel function on $\Theta$,
\begin{equation}
\begin{aligned}
&\phi: \Theta \times \Theta \rightarrow R \\
&P(X\in \Theta)=\frac{\det(\phi(X))}{\det(\phi+I)}\\
\end{aligned}
\end{equation}
{where $\phi(X)$ denotes the kernel matrix and the pairwise $\phi({\bf x}_i,{\bf x}_j)$ is the pairwise correlation between the data ${\bf x}_i$ and ${\bf x}_j$. $\det(\cdot)$ denotes the determinant of matrix.} $I$ is an identity matrix. Since the space $\Theta$ is constant, $\det(\phi+I)$ is a constant value. Therefore, the corresponding diversity prior of transition parameter matrix modeled by DPP can be formulated as
\begin{equation}
  P(X)\propto \det(\phi(X))
\end{equation}
In general, the kernel can be divided into the correlation and the prior part. Therefore, the kernel can be reformulated as
\begin{equation}
\phi({\bf x}_i,{\bf x}_j)=R({\bf x}_i,{\bf x}_j)\sqrt{\pi({\bf x}_i)\pi({\bf x}_j)}
\end{equation}
{where $\pi({\bf x}_i)$ is the prior for the data ${\bf x}_i$ and $R(X)$ denotes the correlation of these data.} These kernels would always induce repulsion between different points and thus a diverse set of points tends to have higher probability. Generally, the vectors are supposed to be uniformly distributed variables. Therefore, the prior $\pi({\bf x}_i)$ is a constant value, and then, the kernel
\begin{equation}
\phi({\bf x}_i,{\bf x}_j)=R({\bf x}_i,{\bf x}_j).
\end{equation}
{The DPPs provide a probability measure over every configuration of subsets on data points.} Based on a similarity matrix over the data and a determinant operator, the DPP assigns higher probabilities to those subsets with dissimilar items. Therefore, it can give lower probabilities to mini-batches which contain the redundant data, and higher probabilities to mini-batches with more diverse data {\cite{36}}. This simultaneously balances the data and generates the stochastic gradients with lower variance. {Moreover, \cite{add_12} further regularizes the DPP (R-DPP) with an arbitrary fixed positive semi-definite matrix inside of the determinant to accelerate the training process.}

{Besides, \cite{add_10} generalizes the diversification of the mini-batch sampling to arbitrary repulsive point processes, such as the Stationary Poisson Disk Sampling (PDS). The PDS is one type of repulsive point process. It can provide point arrangements similar to DPP but with much more efficiency. The PDS indicates that the smallest distance between each pair of sample points should be at least $r$ with respect to some distance measurement $D({\bf x}_i,{\bf x}_j)$ \cite{add_10}, such as the Euclidean distance and the heat kernel. The measurement can be formulated as}

\noindent{{Euclidean distance:}}
\begin{equation}
  D({\bf x}_i,{\bf x}_j)=\|{\bf x}_i-{\bf x}_j\|^2
\end{equation}

\noindent{{Heat kernel:}}
\begin{equation}
  D({\bf x}_i,{\bf x}_j)=e^{\frac{\|{\bf x}_i-{\bf x}_j\|^2}{\sigma}}
\end{equation}
{where $\sigma$ is a positive value.}
{Given a new mini-batch $B$, and the algorithm of PDS can work as follows in each iteration.}
\begin{itemize}
\item {Randomly select a data point ${\bf x}_{new}$.}
\item {If $D({\bf x}_{new},{\bf x}_i) \leq r (\forall {\bf x}_i \in B)$, throw out the point; otherwise add $x_{new}$ in batch $B$.}
\end{itemize}
{The computational complexity of PDS is much lower than that of the DPP.}

{Under these diversification prior, such as the DPP and the PDS, each mini-batch consists of the training samples with more diversity and information,} which can train the model more effectively, and thus the learned model can exact more discriminative features from the objects.

\subsection{Diversification in Active Learning}
{As section \ref{sec:active} shows, active learning can obtain a good performance with less labelled training samples.} However, some selected samples with CTED are similar to each other and contain the overlapping and redundant information. {The highly similar samples make the redundancy of the training samples, and this further decreases the training efficiency, which requires more training samples for a comparable performance.}

{To select more informative and complement samples with the active learning method, some prior works introduce the diversity in the selected samples obtained from CTED (Eq. \ref{03}) \cite{3,1}.}
To promote diversity between the selected samples, \cite{3} enhances CTED with a diversity regularizer
\begin{equation}
\begin{aligned}
&\min\limits_{A,{\bf b}} \|U-UA\|_F^2+\sum_{i=1}^{N_2} \frac{\sum_{1}^{N_2} a_{ij}^2}{b_i}+\alpha \|{\bf b}\|_1 + \gamma{\bf b}^T S {\bf b}\\
&s.t.\ b_i \geq 0, i=1,2,\cdots, {N_2}
\end{aligned}
\end{equation}
where $A=[{\bf a}^1, \cdots,{\bf a}^{N_2}]$, {$\|\cdot\|$ represents the F-norm,} and the similarity matrix $S\in R^{{N_2}\times {N_2}}$ is used to model the pairwise similarities among all the samples, such that larger value of $s_{ij}$ demonstrates the higher similarity between the $i-${th} sample and the $j-${th} one.
Particularly, \cite{3} chooses the cosine similarity measurement to formulate the diversity term. And the diversity term can be formulated as
\begin{equation}\label{eq:87}
s_{ij}=\frac{{\bf a}^i({\bf a}^j)^T}{\|{\bf a}^i\|\|{\bf a}^j\|}.
\end{equation}
As \cite{13} introduces, $s_{ij}$ tends to be zero when ${\bf a}^i$ and ${\bf a}^j$ tends to be uncorrelated.

Similarly, \cite{1} denotes the diversity term in active learning with the angular of the cosine similarity to obtain a diverse set of training samples.
The diversity term can be formulated as
\begin{equation}\label{eq:88}
s_{ij}=\frac{\pi}{2}-\arccos(\frac{{\bf a}^i({\bf a}^j)^T}{\|{\bf a}^i\|\|{\bf a}^j\|}).
\end{equation}
Obviously speaking, when the two vectors become vertical, the vectors tend to be uncorrelated. Therefore, {under the diversification,} the selected samples would be more informative.

{Besides,\cite{add_20} takes advantage of the well-known RBF kernel to measure the diversity of the selected samples, the diversity term can be calculated by}
\begin{equation}\label{eq:86}
s_{ij}=\frac{\|{\bf a}^i-{\bf a}^j\|^2}{\sigma^2}
\end{equation}
{where $\sigma$ is a positive value. Different from Eqs. \ref{eq:87} and Eq. \ref{eq:88} which measure the diversity from the angular view, Eq. \ref{eq:86} calculates the diversity from the distance view. Generally, given two data, if they are similar to each other, the term will have a large value.}

Through adding diversity regularization over the selected samples by active learning, {samples with more information and less redundancy would be chosen for labelling and then used for training.} Therefore, the machine learning process can obtain comparable or {even better performance with limited training samples} than that with {plentiful} training samples.

\subsection{Diversification in Unsupervised Learning}

As subsection \ref{subsec:unsupervised} shows, the unsupervised learning in \cite{gong_ijcnn} is based on the construction of the pseudo classes with the center points. By repulsing the center points from each other, the pseudo classes would be {further enforced to be away from one another}. If we encourage the center points to be diversified and repulse from each other, the learned features from different classes can be more discriminative. Generally, the Euclidean distance can be used to calculate the diversification of the center points. The pseudo label of ${\bf x}_i$ is also calculated by Eq. \ref{eq:un_1}. Then, the unsupervised learning method with the diversity-promoting prior can be formulated as
\begin{equation}\label{eq:un_diversity}
  \max\limits_{W,{\bf c}_i} L(W|U,z_i) + \eta g(W) + \sum_{k=1}^{N_2}\|{\bf c}_{z_k}-{\bf u}_k\| + \gamma \sum_{j\neq k}\|{\bf c}_{j}-{\bf c}_k\|
\end{equation}
where $\gamma$ is a positive value which {denotes} the tradeoff between the optimization term and the diversity term. Under the diversification term, in the training process, the center points {would} be encouraged to repulse from each other. This makes the unsupervised learning process be more effective to obtain discriminative features from samples in different classes.

\section{Model Diversification}\label{sec:model}

{In addition to the data diversification  to improve the performance with more informative and less redundant samples, we can also diversify the model to improve the representational ability of the model directly.} As introduction shows, the machine learning methods aim to learn parameters by the machine itself with the training samples. However, due to the limited and imbalanced training samples, { highly similar  parameters} would be learned by {general} machine learning process. This would lead to the redundancy of the learned model and {negatively affect} the model's representational ability.

Therefore, in addition to the data diversification, one can also diversify the learned parameters in the training process and further improve the representational ability of the model (D-model). Under the diversification prior, each parameter factor can model unique information and the whole factors model a larger proportional of information {\cite{13}}. Another method is to obtain diversified multiple models (D-models) {through  machine learning}. Traditionally, if we train {the} multiple models separately, the obtained representations from different models would be similar and this would lead to the redundancy between different representations. Through regularizing the multiple base models with the diversification prior, different models would be enforced to repulse from each other and each base model can provide choices reflecting multi-modal belief \cite{99}. In the following subsections, we'll introduce the diversity methods for D-model and D-models in detail separately.

\subsection{D-Model}

\begin{figure}
\centering
  \includegraphics[width=0.48\textwidth]{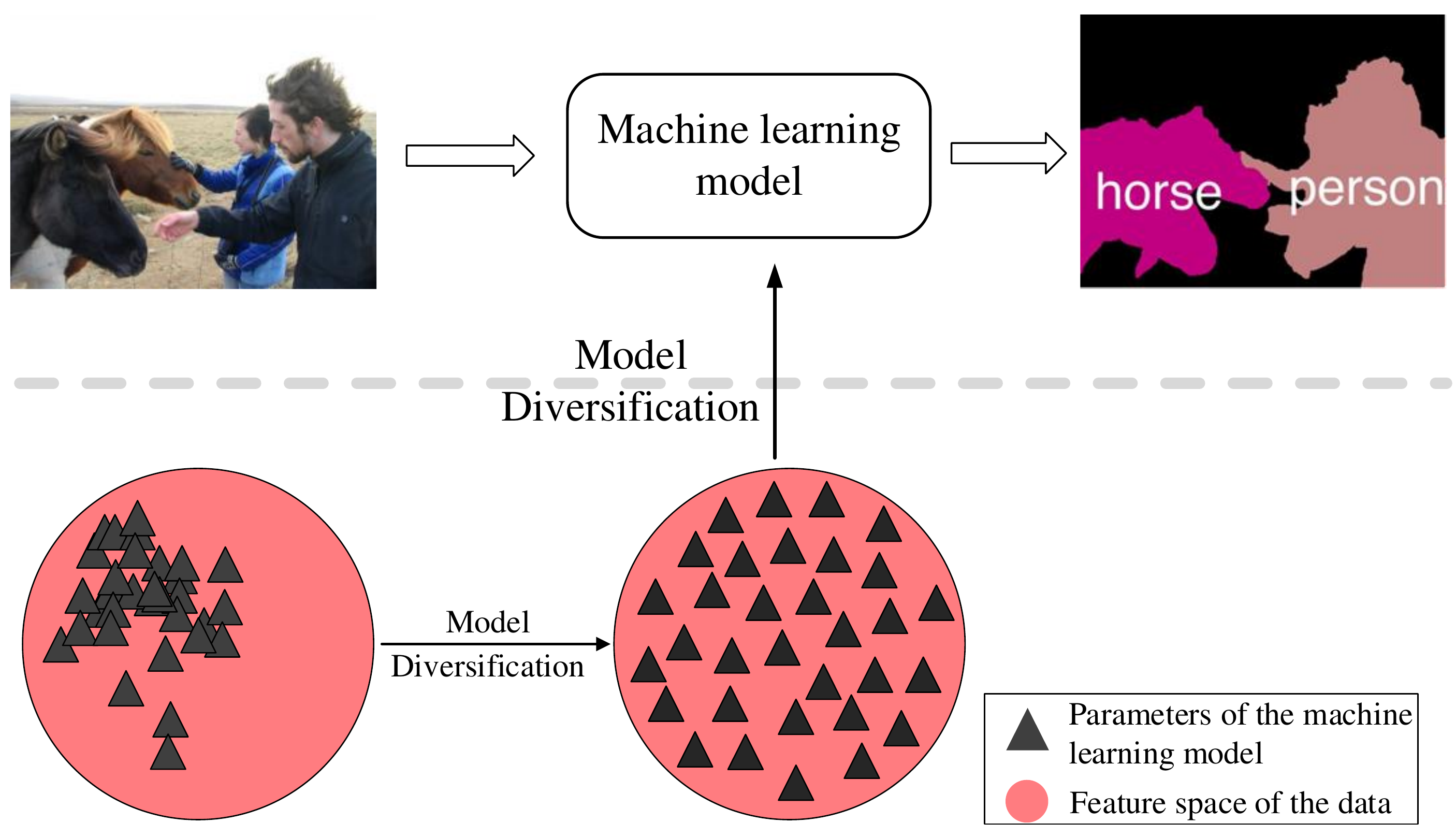}
\caption{Effects of D-model on improving the performance of the machine learning model. Under the model diversification, each parameter factor of the machine learning model tends to model unique information and the whole machine learning model can model more useful information from the objects. Thus, the representational ability can be improved. The figure shows the results from the image segmentation task in \cite{9}. As showed in the figure, the extracted features from the model can better discriminate different objects.}
\label{fig:d_model}       
\end{figure}

The first method tries to diversify the parameters of the model in the training process to directly improve the representational ability of the model.
Fig. \ref{fig:d_model} shows the effects of D-model on improving the performance of the machine learning model. {As Fig. \ref{fig:d_model} shows, under the D-model, each factor would model unique information and the whole factors model a larger proportional of information and then the information will be further improved.}
Traditionally, Bayesian method and posterior regularization method  can be used to impose  diversity over the parameters of  the model. Different diversity-promoting priors have been developed in prior works to measure the diversity between the learned parameter factors according to the special requirements of different tasks. This subsection will mainly introduce the methods which can enforce the diversity of the model and summarize these methods occurred in prior works.

\subsubsection{Bayesian Method}
Traditionally, diversity-promoting priors can be used to measure the diversification of the model. The parameters of the model can be calculated by the Bayesian method as
\begin{equation}
W\propto P(W|X)=P(X|W)\times P(W)
\end{equation}
where $W=[w_1,w_2,\cdots,w_K]$ {denotes} the parameters in the machine learning model, {$K$ is the number of the parameters,} $P(X|W)$ {represents} the likelihood of the training set on the constructed model and $P(W)$ stands for the prior knowledge of the learned model. For the machine learning task at hand, $P(W)$ {describes} the diversity-promoting prior. Then, the machine learning task can be written as
\begin{equation}
W^*=\arg \max\limits_W P(W|X)=\arg\max\limits_W P(X|W)\times P(W)
\end{equation}
The log-likelihood of the optimization can be formulated as
\begin{equation}\label{eq:45}
W^*=\arg\max\limits_W (\log P(X|W)+\log P(W))
\end{equation}

Then, Eq. \ref{eq:45} can be written as the following optimization
\begin{equation}\label{eq:46}
\max\limits_W \log P(X|W)+\log P(W)
\end{equation}
where $\log P(X|W)$ represents the optimization objective of the model, which can be formulated as $L_0$ in subsection \ref{subsec:machine}.
the diversity-promoting prior $\log P(W)$ aims to encourage the learned factors to be diversified. With Eq. \ref{eq:46}, the diversity prior can be imposed over the parameters of the learned model.

\subsubsection{Posterior Regularization Method}

In addition to the former Bayesian method, posterior regularization methods can be also used to impose the diversity property over the learned model \cite{53}.
Generally, the regularization method can add side information into the parameter estimation and thus it can encourage the learned factors to possess {a} specific property. We can also use the posterior regularization to enforce the learned model to be diversified.
The diversity regularized optimization problem can be formulated as
\begin{equation} \label{eq:47}
\max\limits_W L_0 + \gamma f(W)
\end{equation}
where $f(W)$ stands for the diversity {regularization} which measures {the diversity of the factors in the learned model}. $L_0$ represents the optimization term of the model which can be seen in subsection \ref{subsec:machine}. $\gamma$ demonstrates the tradeoff between the optimization and the diversification term.

From Eqs. \ref{eq:46} and \ref{eq:47}, we can find that the posterior regularization has the similar form as the Bayesian method. In general, the optimization (\ref{eq:46}) can be transformed into the form  (\ref{eq:47}). Many methods can be applied to measure the diversity property of the learned parameters. In the following, {we will introduce different diversity priors to realize the D-model in detail.}

\begin{table}
\centering
\caption{Overview of most frequently used diversification method in D-model and the papers in which example measurements can be found.}
\label{table:01}       
\begin{tabular}{p{0.22\textwidth}p{0.2\textwidth}}
\hline\noalign{\smallskip}
Measurements &  Papers   \\
\noalign{\smallskip}\hline\noalign{\smallskip}
Cosine Similarity & \cite{13,14,27,44,46,47,48,49,51,52,59, add_6}  \\
Determinantal Point Process & \cite{10,25,28,29,33,34,35, 104,105,106,107,108,109,111,112,113, add_1}  \\
Submodular Spectral Diversity & \cite{39}  \\
Inner Product & \cite{46,50}  \\
Euclidean Distance & \cite{58,60,61}  \\
Heat Kernel & \cite{54,55,72} \\
Divergence & \cite{58} \\
Uncorrelation and Evenness & \cite{24} \\
$L_{2,1}$ & \cite{31,40,41,42,43,62,63,64,65} \\
\noalign{\smallskip}\hline
\end{tabular}
\end{table}

\subsubsection{Diversity Regularization} \label{subsec:diversity_regularization}
As Fig. \ref{fig:d_model} shows, the diversity regularization encourages the factors to repulse from each other or to be uncorrelated. The key problem with the diversity regularization is the way to calculate the diversification of the factors in the model.
Prior works mainly impose the diversity property into the machine learning process from six aspects, namely the distance, the angular, the eigenvalue, the divergence, the $L_{2,1}$, and the DPP. {The following will introduce the measurements and further discuss the advantages and disadvantages of these measurements.}

{\bf Distance-based measurements.} The simplest way to formulate the diversity between different factors is the Euclidean distance. Generally, {enlarging} the distances between different factors can decrease the similarity between these factors. Therefore, the redundancy between the factors can be decreased and the factors can be diversified. \cite{58, 60, 61} have applied the Euclidean distance as the measurements to encourage the latent factors in machine learning to be diversified.

In general, the larger of the Euclidean distance two vectors have, the more difference the vectors are. {Therefore, we can diversify different vectors through enlarging the pairwise Euclidean distances between these vectors.}
Then,  the diversity regularization by Euclidean distance from Eq. \ref{eq:47} can be formulated as
\begin{equation}\label{eq:85}
f(W)= \sum_{i\neq j}^{K} \|w_i-w_j\|^2
\end{equation}
where $K$ is the number of the factors which we intend to diversify in the machine learning model.
{Since} the Euclidean distance {uses} the distance between different factors to measure the similarity of these factors {, generally the regularizer in Eq. \ref{eq:85} is variant to scale due to the characteristics of the distance.} This may decrease the effectiveness of the diversity measurement and cannot fit for some special models with large scale range.

Another commonly used distance-based method to encourage diversity in the machine learning is the heat kernel \cite{54,55,72}.
The correlation between different factors is {formulated} through Gaussian function and it can be calculated as
\begin{equation}
f(w_i,w_j)=-e^{-\frac{\|w_i-w_j\|^2}{\sigma}}
\end{equation}
where $\sigma$ is a positive value. {The term  measures the correlation between different factors and we can find that when $w_i$ and $w_j$ are dissimilar, $f(w_i,w_j)$ tends to zero.} Then,
{the} diversity-promoting prior by {the} heat kernel from Eq. \ref{eq:46} can be formulated as
\begin{equation}
P(W)=e^{-\gamma \sum_{i\neq j}^{K}e^{-\frac{\|w_i-w_j\|^2}{\sigma}}}
\end{equation}
The {corresponding diversity regularization form} can be formulated as
\begin{equation}
f(W)=-\sum_{i\neq j}^{K}e^{-\frac{\|w_i-w_j\|^2}{\sigma}}
\end{equation}
where $\sigma$ is a positive value.
Heat kernel takes advantage of the distance between {the} factors to encourage the diversity of the model.
{It can be noted that the} heat kernel has the form of Gaussian function and the weight of the diversity penalization is affected by the distance. Thus, the heat kernel presents more variance with the penalization and shows better performance than general Euclidean distance.

All the former distance-based methods encourage the diversity of the model by enforcing the factors away from each other and thus these factors would show more difference. However, it should be noted that the distance-based measurements can be significantly affected by scaling which can limit the performance of the diversity prior over the machine learning.

{\bf Angular-based measurements.} To make the diversity measurement be invariant to scale, some works take advantage of the angular to encourage the diversity of the model. Among these works, the cosine similarity measurement is the {most commonly used} \cite{13, 14}. Obviously, the cosine similarity can measure the similarity between different vectors. In machine learning tasks, it can be used to measure the redundancy between different latent parameter factors \cite{13, 14, 27, 44, 46}. The aim of cosine similarity prior is to encourage different latent factors to be uncorrelated, such that each factor in the learned model can model unique features from the samples.

{The cosine similarity between different factors $w_i$ and $w_j$ can be calculated as \cite{156, add_6}}
\begin{equation}\label{eq:04}
c_{ij}=\frac{<w_i,w_j>}{\|w_i\|\|w_j\|},i\neq j,1\leq i,j\neq K
\end{equation}
{Then,} the diversity-promoting prior of generalized cosine similarity measurement from Eq. \ref{eq:46} can be written as
\begin{equation}\label{eq:12}
P(W)\propto e^{-\gamma(\sum_{i\neq j}c_{ij}^p)^{\frac{1}{p}}}
\end{equation}
It should be noted that when $p$ is set to 1, the diversity-promoting prior over different vectors $w_i(i=1,2,\cdots,K)$ by cosine similarity from Eq. \ref{eq:46} can be formulated as
\begin{equation}\label{eq:05}
P(W)\propto e^{-\gamma \sum_{i\neq j}c_{ij}}
\end{equation}
where $\gamma$ is a positive value.
{It can be noted that under the diversity-promoting prior in Eq. \ref{eq:05}, the
$c_{ij}$ is encouraged to be 0.} Then, $w_i$ and $w_j$ tend to be orthogonal and different factors are encouraged to be uncorrelated and diversified.
Besides, the diversity regularization form {by} the cosine similarity measurement from Eq. \ref{eq:47} can be formulated as
\begin{equation}
f(W)=-\sum_{i\neq j}^{K}\frac{<w_i,w_j>}{\|w_i\|\|w_j\|}
\end{equation}
However, there exist some defects in the former measurement where the measurement is variant to orientation.
To overcome this problem, many works use the angular of cosine similarity to measure the diversity between different factors \cite{26, 44, 47}.

Since the angular between different factors is invariant to translation, rotation, orientation and scale, \cite{26, 44, 47} {develops the angular-based diversifying method for Restricted Boltzmann Machine (RBM)}.
These works use the variance and mean value of the angular between different factors to formulate the diversity of the model to overcome the problem occurred in cosine similarity. The angular between different factors can be formulated as
\begin{equation}
\Gamma_{ij}=\arccos \frac{<w_i,w_j>}{\|w_i\|\|w_j\|}
\end{equation}
Since we do not care about the orientation of the vectors just as \cite{26}, we prefer the angular to be acute or right. From the mathematical view, two factors would tend to be uncorrelated when the angular between the factors enlarges.
Then, the diversity function can be defined as {\cite{add_3, 26, add_7, add_4}}
\begin{equation} \label{eq:84}
f(W)=\Psi(W)-\Pi(W)
\end{equation}
where
\begin{equation*}
\Psi(W)=\frac{1}{K^2}\sum_{i\neq j}\Gamma_{ij},
\end{equation*}
\begin{equation*}
\Pi(W)=\frac{1}{K^2}\sum_{i \neq j}(\Gamma_{ij}-\Psi(W))^2.
\end{equation*}
 {In other words, $\Psi(W)$ denotes the mean of the angular between different factors and $\Pi(W)$ represents the variance of the angular. Generally, a larger $f(W)$ indicates that the weight vectors in $W$ are more diverse. Then, the diversity promoting prior by the angular of cosine similarity measurement can be formulated as}
 \begin{equation}\label{eq:10}
 P(W)\propto e^{\gamma f(W)}
 \end{equation}
{The prior in Eq. \ref{eq:10} encourages the angular between different factors to approach $\displaystyle{\frac{\pi}{2}}$, and thus these factors are enforced to be diversified under the diversification prior. Moreover, the measurement is invariant to scale, translation, rotation, and orientation.}

Another form of the angular-based measurements is to calculate the diversity with the inner product \cite{46,50}.
Different vectors present more diversity when they tend to be more orthogonal. The inner product can measure the orthogonality between different vectors and therefore it can be applied in machine learning models for more diversity.
The general form of diversity-promoting prior by inner product measurement can be written as \cite{46,50}
\begin{equation}
P(W)=e^{-\gamma\sum_{i\neq j}^{K}<w_i, w_j>}.
\end{equation}
Besides, \cite{82} uses the special form of the inner product measurement, which is called exclusivity. The exclusivity between two vectors $w_i$ and $w_j$ is defined as
\begin{equation}
\chi(w_i,w_j)=\|w_i\odot w_j\|_0=\sum_{k=1}^{m} w_i(k)\cdot w_j(k)
\end{equation}
where $\odot$ denotes the Hadamard product, and $\|\cdot\|_0$ denotes the $L_0$ norm. Therefore, the diversity-promoting prior can be written as
\begin{equation}
P(W)=e^{-\gamma\sum_{i\neq j}^{K}\|w_i\odot w_j\|_0}
\end{equation}
Due to the non-convexity and discontinuity of $L_0$ norm, the relaxed exclusivity is calculated as \cite{82}
\begin{equation}
\chi_r(w_i,w_j)=\|w_i\odot w_j\|_1=\sum_{k=1}^{m}|w_i(k)|\cdot |w_j(k)|
\end{equation}
where $\|\cdot\|_1$ denotes the $L_1$ norm. Then, the diversity-promoting prior based on relaxed exclusivity can be calculated as
\begin{equation}\label{eq:89}
P(W)=e^{-\gamma\sum_{i\neq j}^{K}\|w_i\odot w_j\|_1}
\end{equation}
The inner product measurement takes advantage of the characteristics among the vectors and tries to encourage different factors to be orthogonal to enforce the learned factors to be diversified. It should be noted that the measurement can be seen as a special form of cosine similarity measurement. Even though {the inner product measurement} is variant to scale and orientation, in many real-world applications, it is usually considered first to diversify the model since it is easier to implement than other measurements.

Instead of the distance-based and angular-based measurements, the eigenvalues of the kernel matrix can also be used to encourage different factors to be orthogonal and diversified.
Recall that, for an orthogonal matrix, all the eigenvalues of the kernel matrix are equal to 1. Here, we denote $\kappa(W)=WW^T$ as the kernel matrix of $W$. Therefore, when we constrain the eigenvalues to 1, the obtained vectors would tend to be orthogonal {\cite{add_2, add_9}}. {Three ways are generally used to encourage the eigenvalues to approach constant 1, including the submodular spectral diversity (SSD) measurement, the uncorrelation and evenness measurement, and the log-determinant divergence (LDD).} In the following, the two form of the eigenvalue-based measurements will be introduced in detail.

{\bf Eigenvalue-based measurements.} {As the former denotes,} $\kappa(W)=WW^T$ stands for the kernel matrix of the latent factors. Two commonly used methods to promote diversity in the machine learning process based on the kernel matrix would be introduced. The first method is the {submodular spectral diversity (SSD)}, which is based on the eigenvalues of the kernel matrix. \cite{39} introduces the {SSD measurement} in the process of feature selection, which aims to select a diverse set of features. Feature selection is a key component in many machine learning settings. The process involves choosing a small subset of features in order to build a model to approximate the target concept well.

The {SSD measurement} uses the square distance to encourage the eigenvalues to approach 1 directly. Define $(\lambda_1,\lambda_2,\cdots,\lambda_K)$ as the eigenvalues of the kernel matrix. Then,
the diversity-promoting prior by SSD from Eq. \ref{eq:46} can be formulated as \cite{39}
\begin{equation}
P(W)=e^{-\gamma\sum_{i=1}^{K}(\lambda_i(\kappa(W))-1)^2}
\end{equation}
where $\gamma$ is also a positive value.
From Eq. \ref{eq:47}, {the diversity regularization $f(W)$ can be formulated as}
\begin{equation}
f(W)=-\sum_{i=1}^{K}(\lambda_i(\kappa(W))-1)^2
\end{equation}
This measurement regularizes the variance of the eigenvalues of the matrix. Since all the eigenvalues are enforced to approach 1, the obtained factors tend to be more orthogonal and thus the model can present more diversity.

Another diversity measurement based on the kernel matrix is the uncorrelation and evenness \cite{24}. This measurement encourages
the learned factors to be uncorrelated and to play equally important roles in modeling data. Formally, this amounts to encouraging the kernel matrix of the vectors to have more uniform eigenvalues.
The basic idea is to normalize the eigenvalues into a probability simplex and encourage the discrete distribution parameterized by the normalized eigenvalues to have small Kullback-Leibler (KL) divergence with the uniform distribution \cite{24}. Then, the diversity-promoting prior by uniform eigenvalues from Eq. \ref{eq:46} is formulated as
\begin{equation}
P(W)=e^{-\gamma(\frac{tr((\frac{1}{d}\kappa(W))\log(\frac{1}{d}\kappa(W)))}{tr(\frac{1}{d}\kappa(W))}-\log tr(\frac{1}{d}\kappa(W)))}
\end{equation}
subject to $\kappa(W)\succ 0$ ( $\kappa(W)$ is positive definite matrix) and $W{\bf 1}=0$, where $\kappa(W)$ is the kernel matrix.
Besides, the diversity-promoting uniform eigenvalue regularizer (UER) from Eq. \ref{eq:47} is formulated as
\begin{equation}
f(W)=-[\frac{tr((\frac{1}{d}\kappa(W))\log(\frac{1}{d}\kappa(W)))}{tr(\frac{1}{d}\kappa(W))}-\log tr(\frac{1}{d}\kappa(W))]
\end{equation}
where $d$ is the dimension of each factor.

{Besides, \cite{add_9} takes advantage of the log-determinant divergence (LDD) to measure the similarity between different factors. The diversity-promoting prior in \cite{add_9} combines the orthogonality-promoting LDD regularizer with the sparsity-promoting $L_1$ regularizer. Then, the diversity-promoting prior from Eq. \ref{eq:46} can be formulated as}
\begin{equation}
P(W)=e^{-\gamma(tr(\kappa(W))-\log\det(\kappa(W))+\tau |W|_1)}
\end{equation}
{where $tr(\cdot)$ denotes the matrix trace. Then, the corresponding regularizer from Eq. \ref{eq:47} is formulated as}
\begin{equation}
f(W)=-(tr(\kappa(W))-\log\det(\kappa(W))+\tau |W|_1)).
\end{equation}
{The LDD-based regularizer can effectively promote nonoverlap \cite{add_9}. Under the regularizer, the factors would be sparse and orthogonal simultaneously. }

These eigenvalue-based measurements calculate the diversity of the factors from the kernel matrix view. They not only consider the pairwise correlation between the factors, but also take the multiple correlation into consideration. Therefore, they generally present better performance  than the distance-based and angular-based methods which only consider the pairwise correlation. However, the eigenvalue-based measurements would cost more computational sources in the implementation. Moreover, the gradient of the diversity term which is used for back propagation would be complex to compute and usually requires special processing methods, such as projected gradient descent algorithm \cite{24} for the uncorrelation and evenness.

{\bf DPP measurement.}
Instead of the eigenvalue-based measurements, another measurement which takes the multiple correlation into consideration is the determinantal point process (DPP) measurement.
{As subsection \ref{subsec:dpp} shows, the DPP on the parameter factors $W$ has the form as}
\begin{equation}
P(W)\propto \det(\phi(W)).
\end{equation}
{Generally, it can encourage the learned factors to repulse from each other.} Therefore, the DPP-based diversifying prior can obtain machine learning models with a diverse set of the learned factors other than a redundant one. Some works have shown that the DPP prior is usually not arbitrarily strong for some special case when applied into machine learning models \cite{37}. To encourage the DPP prior strong enough for all the training data, the DPP prior is augmented by an additional positive parameter $\gamma$. Therefore, just as section \ref{subsec:dpp}, the DPP prior can be reformulated as
\begin{equation}
P(W)\propto \det(\phi(W))^\gamma
\end{equation}
where $\phi(W)$ denotes the kernel matrix and $\phi(w_i, w_j)$ demonstrates the pairwise correlation between $w_i$ and $w_j$.
The learned factors are usually normalized, and thus the optimization for machine learning can be written as
\begin{equation}
\max\limits_{W}\log P(X|W)+\gamma \log(\det(\phi(W)))
\end{equation}
where $f(W)=\log(\det(\phi(W)))$ represents the diversity term for machine learning. It should be noted that different kernels can be selected according to the special requirements of different machine learning tasks \cite{38, gong_cnn}. For example, in \cite{gong_cnn}, the similarity kernel is adopted for the DPP prior which can be formulated as
\begin{equation}
\phi(w_i, w_j)=\frac{<w_i, w_j>}{\|w_i\|\|w_j\|}.
\end{equation}
{When we set the cosine similarity as the correlation kernel $\phi$, from geometric interpretation, the DPP prior $P(W)\propto \det(\phi(W))$ can be seen as the volume of the parallelepiped spanned by the columns of $W$ \cite{34}. Therefore, diverse sets are more probable because their feature vectors are more orthogonal, and hence span larger volumes.}
{It should be noted that most of the diversity} measurements consider the pairwise correlation between the factors and ignore the multiple correlation between three or more factors. {While the DPP measurement} takes advantage of the merits of the DPP to make use of the multiple correlation by calculating the similarity between multiple factors.

{\bf $L_{2,1}$ measurement.} While all the former measurements promote the diversity of the model from the pairwise or multiple correlation view, many prior works prefer to use the $L_{2,1}$ for diversity since $L_{2,1}$ can take advantage of the group-wise correlation and obtain a group-wise sparse representation of the  latent factors $W$ {\cite{31, 40, 41, add_5}}.

It is well known that the  $L_{2,1}$-norm leads to the group-wise sparse representation of $W$. $L_{2,1}$ can also be used to measure the correlation between different parameter factors and diversify the learned factors to improve the representational ability of the model. Then, the $L_{2,1}$ prior from Eq. \ref{eq:46} can be calculated as
\begin{equation}
P(W)=e^{-\gamma\sum_{i}^{K}(\sum_{j}^{n}|w_i(j)|)^2}
\end{equation}
where $w_i(j)$ means the $j-$th entry of $w_i$.
The internal $L_1$ norm encourages different factors to be sparse, while the external $L_2$ norm is used to control the complexity of entire model.
Besides, the diversity term based on $f(W)$ from Eq. \ref{eq:47} can be formulated as
\begin{equation}
f(W)=-\sum_{i}^{K}(\sum_{j}^{n}|w_{i}{(j)}|)^2
\end{equation}
where $n$ is the dimension of each factor $w_i$.
The internal $L_1-$norm encourages different factors to be sparse, while the external $L_2-$norm is used to control the complexity of entire model.

In most of the machine learning models, the parameters of the model can be looked as the vectors and diversity of these factors can be calculated from {the mathematical view just as these former measurements.} When the norm of the vectors are constrained to constant 1, we can also take these factors as the probability distribution. Then, the diversity between the factors can be also measured from the Bayesian view.

{\bf Divergence measurement.} Traditionally, divergence, which is generally used Bayesian method to measure the difference between different distributions, can be used to promote  diversity of the learned model \cite{58}.

Each factor is processed as a probability distribution firstly. Then, the divergence between factors $w_i$ and $w_j$ can be calculated as
\begin{equation}
D(w_i \| w_j)=\sum_{k=1}^n(w_{i}{(k)}\log\frac{w_{i}{(k)}}{w_{j}{(k)}}-w_{i}{(k)}+w_{j}{(k)})
\end{equation}
subject to $\|w_i\|=1$.

The divergence can measure the dissimilarity between the learned factors, such that the diversity-promoting regularization by divergence from Eq. \ref{eq:47} can be formulated as \cite{58}
\begin{equation}
\begin{aligned}
f(W)=&\sum_{i\neq j}^{K}D(w_i \| w_j)  \\
=&\sum_{i\neq j}^{K}\sum_{k=1}^{n}(w_{i}{(k)}\log \frac{w_{i}{(k)}}{w_{j}{(k)}}-w_{i}{(k)}+w_{j}{(k)})
\end{aligned}
\end{equation}
 The measurement takes advantage of the characteristics of the divergence to measure the dissimilarity between different distributions. However, the norm of the learned factors need to satisfy $\|w_i\|=1$ which limits the application field of the diversity measurement.

In conclusion, there are numerous approaches to diversify the learned factors {in machine learning models.} A summary of the most frequently encountered diversity methods is shown in Table \ref{table:01}. Although most papers use slightly different specifications for their diversification of the learned model, the fundamental representation of the diversification is similar. It should also be noted that the thing in common among the studied diversity methods is that the diversity enforced in a pairwise form between members strikes a good balance between complexity and effectiveness \cite{82}. In addition, different applications should choose the proper diversity measurements according to the specific requirements of different machine learning tasks.

\subsubsection{Analysis}\label{subsec:analysis}
These diversity measurements can calculate the similarity between different vectors and thus encourage the diversity of the machine learning model. However, there exists the difference between these measurements. The details of these diversity measurements can be seen in Table \ref{table:comparison}. It can be noted from the table that all these methods take advantage of the pairwise correlation except $L_{2,1}$ which uses the group-wise correlation between different factors. Moreover, the determinantal point process, submodular spectral diversity, and uncorrelation and evenness can also take advantage of correlation among three or more factors.

Another property of these diversity measurement is scale invariant. Scale invariant can make the diversity of the model be invariant w.r.t. the norm of these factors. The cosine similarity measurement calculates the diversity via the angular between different vectors. As a special case for DPP, the cosine similarity can be used as the correlation term $R(w_i,w_j)$ in DPP and thus the DPP measurement is scale invariant. Besides, for divergence measurement, since the factors are constrained with $\|w_i\|=1$, the measurement is scale invariant.

\begin{table*}
\centering
\caption{Comparisons of Different Measurements. $\bigcirc$ represents that the measurement possess the property while $\times$ means the measurement does not possess the property.}
\label{table:comparison}       
\setlength{\tabcolsep}{3pt}
\begin{tabular}{p{0.2\textwidth}p{0.14\textwidth}p{0.14\textwidth}p{0.16\textwidth}p{0.14\textwidth}p{0.0\textwidth}}
\hline\noalign{\smallskip}
Measurements &  \centering{Pairwise Correlation} & \centering{Multiple Correlation} & \centering{Group-wise Correlation} & \centering{{Scale Invariant}} &  \\
\noalign{\smallskip}\hline\noalign{\smallskip}
Cosine Similarity & \centering{$\bigcirc$} & \centering{$\times$} & \centering{$\times$} & \centering{{$\bigcirc$}} & \\
Determinantal Point Process & \centering{$\bigcirc$} & \centering{$\bigcirc$} &\centering{$\times$} & \centering{$\bigcirc$} & \\
Submodular Spectral Diversity &\centering{$\bigcirc$} &\centering{$\bigcirc$} & \centering{$\times$}& \centering{$\bigcirc$} & \\
Euclidean Distance &\centering{$\bigcirc$} & \centering{$\times$}&\centering{$\times$} & \centering{$\times$} & \\
Heat Kernel & \centering{$\bigcirc$}&\centering{$\times$} & \centering{$\times$}&\centering{$\times$} &  \\
Divergence &\centering{$\bigcirc$} & \centering{$\times$}&\centering{$\times$} &\centering{$\bigcirc$} & \\
Uncorrelation and Evenness &\centering{$\bigcirc$} & \centering{$\bigcirc$}& \centering{$\times$}& \centering{$\bigcirc$}& \\
$L_{2,1}$ &\centering{$\times$} & \centering{$\times$}& \centering{$\bigcirc$}&\centering{$\times$} & \\
Inner Product &\centering{$\bigcirc$} &\centering{$\times$} & \centering{$\times$}& \centering{$\times$}& \\
\noalign{\smallskip}\hline
\end{tabular}
\end{table*}

These measurements can encourage diversity within different vectors. Generally, the machine learning models can be looked as the set of latent parameter factors, which can be represented as the vectors. These factors can be learned and used to represent the objects. In the following, we'll mainly summarize the methods to diversify the ensemble learning (D-models) for better  performance of machine learning tasks.

\subsection{D-Models}\label{subsec:d_models}
\begin{figure}
\centering
  \includegraphics[width=0.48\textwidth]{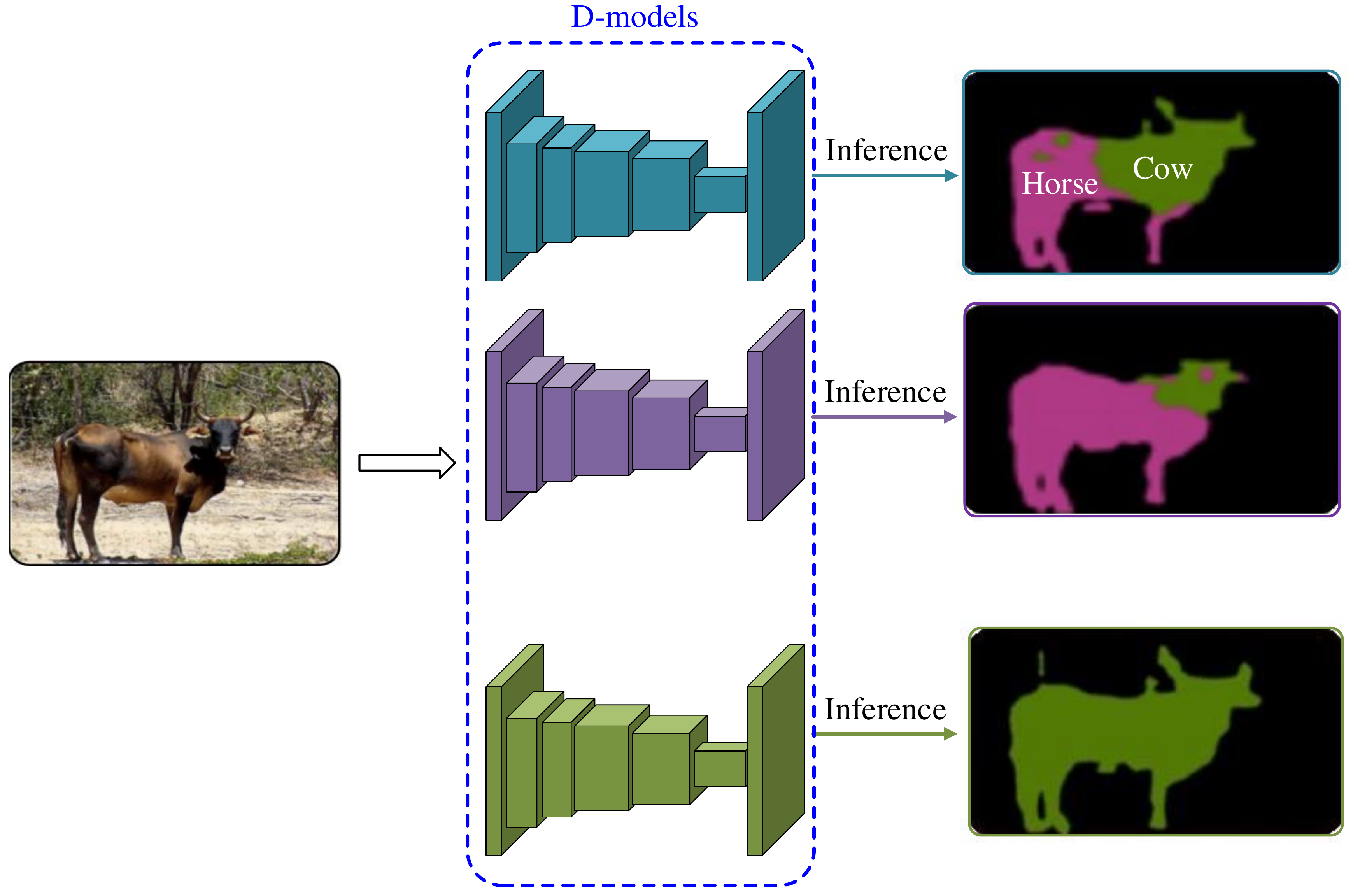}
\caption{Effects of D-models for improving the performance of the machine learning model.  The figure shows the image segmentation task from the prior work \cite{99}. The single model often produce solutions with low expected loss and step into the sub-optimal results. Besides, general ensemble learning usually provide multiple choices with great similarity. Therefore, this work summarizes the methods which can diversify the ensemble learning (D-models). As the figure shows, under the model diversification, each model of the ensemble can produce different outputs reflecting multi-modal belief \cite{99}.}
\label{fig:d_models}       
\end{figure}

The former subsection introduces the way to diversify the parameters in single model and improve the representational ability of the model directly. Much efforts have been done to obtain the highest probability configuration of the machine learning models in prior works. However, even when the training samples are sufficient, the maximum a $posteriori$ (MAP) solution could also be sub-optimal.
In many situations, one could benefit from additional representations with multiple models. As Fig. \ref{fig:03} shows, ensemble learning (the way for training multiple models) has already occurred in many prior works. However, traditional ensemble learning methods to train multiple models may provide representations that tend to be similar while  the representations obtained from different models are desired to provide complement information.
Recently, many diversifying methods have been proposed to overcome this problem. As Fig. \ref{fig:d_models} shows, under the model diversification, each base model of the ensemble can produce different outputs reflecting multi-modal belief. Therefore, the whole performance of the machine learning model can be improved. Especially, the D-models play an important role in structured prediction problems with multiple reasonable interpretations, of which only one is the groundtruth \cite{99}.

{Denote $W_i (i=1,2,\cdots, s)$ and $P(W_i)$ as the parameters and the inference from the $i$th model where $s$ is the number of the parallel base models.}
Then, the optimization of the machine learning to obtain multiple models can be written as
\begin{equation}
\max\limits_{W_1, W_2, \cdots, W_s} \sum_{i=1}^s L(W_i|X_i)
\end{equation}
where $L(W_i|X_i)$ represents the optimization term of the $i${th} model and $X_i$ denotes the training samples of the $i$th model. Traditionally, the training samples are randomly divided into multiple subsets and each subset trains a corresponding model. However, selecting subsets randomly may lead to the redundancy between different representations. Therefore, the first way to obtain multiple diversified models is to diversify these training samples over different base models, which we call sample-based methods.

Another way to encourage the diversification between different models is to measure the similarity between different base models with a special similarity measurement and encourage different base models to be diversified in the training process, which is summarized as the optimization-based methods. The optimization of these methods can be written as
\begin{equation}\label{eq:58}
\max\limits_{W_1, W_2, \cdots, W_s} \sum_{i=1}^s L(W_i|X)+\gamma\Gamma(W_1, W_2, \cdots, W_s)
\end{equation}
where $\Gamma(W_1, W_2, \cdots, W_s)$ measures the diversification between different base models. These methods are similar to the methods for D-model in former subsection.

Finally, some other methods try to obtain large amounts of models and select the top-$L$ models as the final ensemble, which is called the ranking-based methods in this work. In the following, we'll summarize different methods for diversifying multiple models from the three aspects in detail.

\begin{table}
\centering
\caption{{Overview of most frequently used diversification method in D-models and the papers in which example measurements can be found.}}
\label{table:02}       
\begin{tabular}{l | p{0.12\textwidth}p{0.15\textwidth}}
\hline\noalign{\smallskip}
Methods & Measurements & Papers   \\
\noalign{\smallskip}\hline
\multirow{4}*{{Optimization-based}} & Divergence & \cite{118,114}  \\
\cline{2-3}

&Renyi-entropy & \cite{77}  \\
\cline{2-3}

& Cross Entropy & \cite{74,80}  \\
\cline{2-3}
& Cosine Similarity & \cite{117, 82}  \\
\cline{2-3}
& $L_{2,1}$ & \cite{41}  \\
\cline{2-3}
& NCL & \cite{115,116, 92, 129}  \\
\cline{2-3}
& Others & \cite{90, 118, 119, 30, 117, 120, 121}  \\
\hline
{Sample-based} & \centering{-} & \cite{15,75,88,99,104} \\
\hline
{Ranking-based} & \centering{-} & \cite{85, 22, 105} \\
\hline
\end{tabular}
\end{table}

\subsubsection{Optimization-Based Methods}\label{subsubsec:optimization}

Optimization-based methods are one of the most commonly used methods to diversify multiple models. These methods try to obtain multiple diversified models by optimizing a given objective function as Eq. \ref{eq:58} shows, which includes a diversity measurement. Just as the diversity of D-model in prior subsection, the main problem of these methods is to define diversity measurements which can calculate the difference between different models.

{Many prior works \cite{30, 90, 117, 84, add_14, add_3} have summarized some pairwise diversity measurements, such as  Q-statistics measure \cite{118, add_3}, correlation coefficient measure \cite{118, add_3}, disagreement measure \cite{119, 90, 113}, double-fault measure \cite{120, 90, 113}, $k$ statistic measure \cite{121}, Kohavi-Wolpert variance \cite{30, 113}, inter-rater agreement \cite{30, 113}, the generalized diversity \cite{30} and the measure of "Difficult" \cite{30, 113}.} Recently, some more measurements have also been developed, including not only the pairwise diversity measurement \cite{114, 118, 117} but also the measurements which calculate the multiple correlation and others \cite{41,78,92, 116}. This subsection will summarize these methods systematically.

{\bf Bayesian-based measurements.} Similar to D-model, Bayesian methods can also be applied in D-models. Among these Bayesian methods, divergence is the most commonly used one. As former subsection shows, the divergence can measure the difference between different distributions. The way to formulate the diversity-promoting term by the divergence method over the ensemble learning is to calculate the divergence between different distributions from the inference of different models, respectively \cite{114,118}. The diversity-promoting term by divergence from Eq. \ref{eq:58} can be formulated as
\begin{equation}
\begin{aligned}
\Gamma(W_1, W_2, \cdots, & W_s)=  \\
\sum_{i,j}^{s}\sum_{k=1}^{n}(P(W_{i}&{(k)})\log\frac{P(W_{i}{(k)})}{P(W_{j}{(k)})}-P(W_{i}{(k)})+P(W_{j}{(k)}))
\end{aligned}
\end{equation}
where $W_i{(k)}$ represents the $k-$th entry in $W_i$. $P(w_i)$ denotes the distributions of the inference from the $i-${th} model. The former diversity term can increase the difference between the inference obtained from different models and would encourage the learned multiple models to be diversified.

In addition to the divergence measurements, Renyi-entropy which measures the kernelized distances between the images of samples and the center of ensemble in the high-dimensional feature space can also be used to encourage the diversity of the learned multiple models \cite{77}.
The Renyi-entropy is calculated based on the Gaussian kernel function and the diversity-promoting term from Eq. \ref{eq:58} can be formulated as
\begin{equation}
\begin{aligned}
\Gamma(W_1, W_2, \cdots, W_s)&= \\
-\log[\frac{1}{s^2}\sum_{i=1}^{s}\sum_{j=1}^{s}&G(P(W_i)-P(W_j),2\sigma^2)]
\end{aligned}
\end{equation}
where $\sigma$ is a positive value and $G(\cdot)$ represents the Gaussian kernel function, which can be calculated as
\begin{equation}
\begin{aligned}
G(W_i-W_j,&2\sigma^2)=  \\
\frac{1}{(2\pi)^{\frac{d}{2}}\sigma^d}&\exp\{-\frac{(P(W_i)-P(W_j))^T(P(W_i)-P(W_j))}{2\sigma^2}\}
\end{aligned}
\end{equation}
where $d$ denotes the dimension of $W_i$. Compared with the divergence measurement, the Renyi-entropy measurement can be more fit for {the machine learning model} since the difference can be adapted for different models with different value  $\sigma$. However, the Renyi-entropy would cost more computational sources and the update of the ensemble would be more complex.

Another measurement which is based on the Bayesian method is the cross entropy measurement\cite{74,80,93}. The cross entropy measurement uses the cross entropy between pairwise distributions to encourage two distributions to be dissimilar and then different base models could provide more complement information.
 Therefore, the cross-entropy between different base models can be calculated as
\begin{equation}
\begin{aligned}
\Gamma(w_i,w_j)=&\frac{1}{n}\sum_{k=1}^{n}(P_k(w_i)\log P_k(w_j)  \\
&+(1-P_k(w_i))\log (1-P_k(w_j)))
\end{aligned}
\end{equation}
where $P(W_i)$ is the inference of the $i-$th model and $P_k(W_i)$ is the probability of the sample belonging to the $k$th class.
According to the characteristics of the cross entropy and the requirement of the diversity regularization, the diversity-promoting regularization of the cross entropy from Eq. \ref{eq:58} can be formulated as
\begin{equation}
\begin{aligned}
\Gamma(w_1,w_2,\cdots,w_K)=&\frac{1}{n}\sum_{i,j}^{s}\sum_{k=1}^{n}(P_k(w_i)\log P_k(w_j) \\
&+(1-P_k(w_i))\log(1-P_k(w_j)))
\end{aligned}
\end{equation}
We all know that the larger the cross entropy is, the more difference the distributions are. Therefore, under the cross entropy measurement, different models can be diversified and provide more complement information.
Most of the former Bayesian methods promote the diversity in the learned {multiple base models} by calculating the pairwise difference between these base models. However, these methods ignore the correlation among three or more base models.

To overcome this problem, \cite{78} proposes a hierarchical pair competition-based parallel genetic algorithm (HFC-PGA) to increase the diversity among the component neural networks. The HFC-PGA takes advantage of the average of all the distributions from the ensemble to calculate the difference of each base model. The diversity term by HFC-PGA from Eq. \ref{eq:58} can be formulated as
\begin{equation}
\Gamma(W_1,W_2,\cdots,W_s)=\sum_{j=1}^{s}(\frac{1}{s}\sum_{i=1}^{s}P(W_i)-P(W_j))^2
\end{equation}
It should be noted that the HFC-PGA takes advantage of multiple correlation between the multiple models. However, the HFC-PGA method uses the fix weight to calculate the mean of the distributions and further calculate the covariance of the multiple models which usually cannot fit for different tasks. This would limit the performance of the diversity promoting prior.

To deal with the {shortcomings} of the HFC-PGA, negative correlation learning (NCL) tries to reduce the covariance among all the models while the variance and bias terms are not increased \cite{115,116, 92, 129}. The NCL trains the base models simultaneously in a cooperative manner that decorrelates individual errors. The penalty term can be designed in different ways depending on whether the models are trained sequentially or parallelly.
\cite{115} uses the penalty to decorrelate the current learning model with all previously learned models
\begin{equation}
\Gamma(W_1,W_2,\cdots,W_s)=\sum_{k=1}^{s}(P(W_k)-l)\sum_{j=1}^{k-1}(P(W_j)-l)
\end{equation}
where $l$ represents the target function which is a desired output scalar vector.
{Besides, define $\bar{P}=\sum_{i=1}^{s}\alpha_i P(W_i)$ where $\sum_{i=1}^{s}\alpha_i=1$.} Then, the penalty term can also be defined to reduce the correlation mutually among all the learned models by using the actual distribution $\bar{P}$ obtained from each model instead of the target function $l$ {\cite{116,92,add_14}}.
\begin{equation}
\Gamma(W_1,W_2,\cdots,W_s)=\sum_{k=1}^{s}(P(W_k)-\overline{P})\sum_{j=1}^{k-1}(P(W_j)-\overline{P})
\end{equation}
This measurement uses the covariance of the inference results obtained from the multiple models to reduce the correlation mutually among the learned models. Therefore, the learned multiple models can be diversified.
{In addition, \cite{76} further combines the NCL with sparsity.} The sparsity is purely pursued by the $L_1$ norm regularization without considering the complementary characteristics of the available base models.

Most of the Bayesian methods promote diversity in ensemble learning mainly by increasing the difference between the probability distributions of the inference of different base models. There exist other methods which can promote diversity over the parameters of each base model directly.

{\bf Cosine similarity measurement.} Different from the Bayesian methods which promote diversity from the distribution view,
\cite{117} introduces the cosine similarity measurements to calculate the difference between different models from geometric view. Generally, the diversity-promoting term from Eq. \ref{eq:58} can be written as
\begin{equation}
\Gamma(W_1, W_2, \cdots, W_s)=-\sum_{i\neq j}^{s}\frac{<W_i,W_j>}{\|W_i\|\|W_j\|}.
\end{equation}
In addition, as a special form of angular-based measurement,  a special form of inner product measurement, termed as exclusivity, has been proposed by \cite{82} to obtain diversified models. It can jointly suppress the training error of ensemble and enhance the diversity between bases. The diversity-promoting term by exclusivity (see Eq. \ref{eq:89} for details) from Eq. \ref{eq:58} can be written as
\begin{equation}
\Gamma(W_1, W_2, \cdots, W_s)=-\sum_{i\neq j}^{s}\|W_i\bigodot W_j\|_1
\end{equation}
These measurements try to encourage the pairwise models to be uncorrelated such that each base model can provide more complement information.

{\bf $L_{2,1}$ measurement.} Just as the former subsection, $L_{2,1}$ norm can also be used as the diversification of multiple models\cite{41}. the diversity-promoting regularization by $L_{2,1}$ from Eq. \ref{eq:58} can be formulated as
\begin{equation}
\Gamma(W_1, W_2, \cdots, W_s)=-\sum_{i}^{s}(\sum_{j}^{K}|W_i(j)|)^2
\end{equation}
The $L_{2,1}$ measurement uses the group-wise correlation between different base models and favors selecting diverse models residing in more groups.

Some other diversity measurements have been proposed for deep ensemble.
\cite{91} reveals that it may be better to ensemble many instead of all of the neural networks at hand. The paper develops an approach named Genetic Algorithm
based Selective Ensemble (GASEN) to obtain different weights of each neural network. Then, based on the obtained weights, the deep ensemble can be formulated. Moreover, \cite{79} also encourages the diversity of the deep ensemble by defining a pair-wise similarity between different terms.

{These optimization-based methods utilize the correlation between different models and try to repulse these models from one another. The aim is to enforce these representations which are obtained from different models to be diversified and thus these base models can provide outputs reflecting multi-modal belief.}

\subsubsection{Sample-Based Methods}\label{subsubsec:sample}
In addition to diversify the ensemble learning from the optimization view, we can also diversify the models from the sample view. {Generally, we randomly divide the training set into multiple subsets where each base model corresponds to a specific subset which is used as the training samples.} However, there exists the overlapping between the representations of different base models. This may cause the redundancy and even decrease the performance of the ensemble learning due to the reduction of the training samples over each model by the division of the whole training set.

To overcome this problem and provide more complement information from different models, \cite{99} develops a novel method by dividing the training samples into multiple subsets by assigning the different training samples into the specified subset where the corresponding learned model shows the lowest predict error. Therefore, each base model would focus on modeling the features from specific classes. Besides, clustering is another popular method to divide the training samples for different models \cite{104}. Although diversifying the obtained multiple subsets can make the multiple models provide more complement information, the less of training samples by dividing the whole training set will show negative effects over the performance.

To overcome this problem, another way to enforce different models to be diversified is to assign each sample with a specified weight  \cite{15}. By training different base models with different weights of samples, each base model can focus on complement information from the samples. The detailed steps in \cite{15} are as follows:
\begin{itemize}
\item {Define the weights over each training sample randomly, and train the model with the given weights;}
\item {Revise the weights over each training sample based on the final loss from the obtained model, and train the second model with the updated weights;}
\item {Train $M$ models with the aforementioned strategies.}
\end{itemize}

The former methods take advantage of the labelled training samples to enforce the diversity of multiple models. There exists another method, namely Unlabeled Data to Enhance Ensemble (UDEED) \cite{75}, which {focuses on} the unlabelled samples to promote diversity of the model. Unlike the existing semi-supervised ensemble methods where error-prone pseudo-labels are estimated for unlabelled data to enlarge the labelled data to improve accuracy. UDEED works by maximizing accuracies of base models on labelled data while maximizing diversity among them on unlabelled data.
Besides, \cite{88} combines the different initializations, different training sets and different feature subsets to encourage the diversity of the multiple models.

The methods in this subsection process on the training sets to diversify different models. By training different models with different training samples or samples with different weights, these models would provide different information and thus the whole models could provide a larger proportional of information.

\subsubsection{Ranking-Based Methods}\label{subsubsec:ranking}
Another kind of methods to promote diversity in the obtained multiple models is ranking-based methods.
All the models is first ranked according to some criterion, and then the top-$L$ are selected to form the final ensemble. Here, \cite{85} focuses on pruning techniques based on forward/backward selection, since they allow a direct comparison with the simple estimation of accuracy from different models.

Cluster can be also used as ranking-based method to enforce diversity of the multiple models \cite{105}. In \cite{105}, each model is first clustered based on the similarity of their predictions, and then each cluster is then pruned to remove redundant models, and finally the remaining models in each cluster are finally combined as the base models.

In addition to the former mentioned methods, \cite{22} provides multiple diversified models by selecting different sets of multiple features. Through multi-scale or other tricks, each sample will provide large amounts of features, and then choose top-$L$ multiple features from the all the features as the base features (see \cite{22} for details). Then, each base feature from the samples is used to train a specific model, and the final inference can be obtained through the combination of these models.

In summary, this paper summarizes the diversification methods for D-models from three aspects: optimization-based methods, sample-based methods, and ranking-based methods. The details of the most frequently encountered diversity methods is shown in Table \ref{table:02}. Optimization-based methods encourage the multiple models to be diversified by imposing diversity regularization  between different base models while optimizing these models. In contrary, sample-based methods mainly obtain diversified models by training different models with specific training sets. Most of the prior works focus on diversifying the ensemble learning from the two aspects. While the ranking-based methods try to obtain the multiple diversified models by choosing the top-$L$ models.
{The researchers can choose the specific method for D-models based on the special requirements of the machine learning tasks.}

\section{Inference Diversification}\label{sec:inference}
The former section summarizes the methods to diversify different parameters in the model or multiple base models. The D-model focuses on the diversification of parameters in the model and improves the representational ability of the model itself while D-models tries to obtain multiple diversified base models, each of which focus on modeling different features from the samples. These works improve the performance of the machine learning process {in the modeling stage (see Fig. \ref{fig:01} for details).}
In addition to these methods, there exist some other works focusing on obtaining multiple choices in the inference of the machine learning model.
{This section will summarize these diversifying methods in the inference stage.}  To introduce the methods for inference diversification in detail, we choose the graph model as the representation of the machine learning models.

\begin{figure}
\centering
  \includegraphics[width=0.48\textwidth]{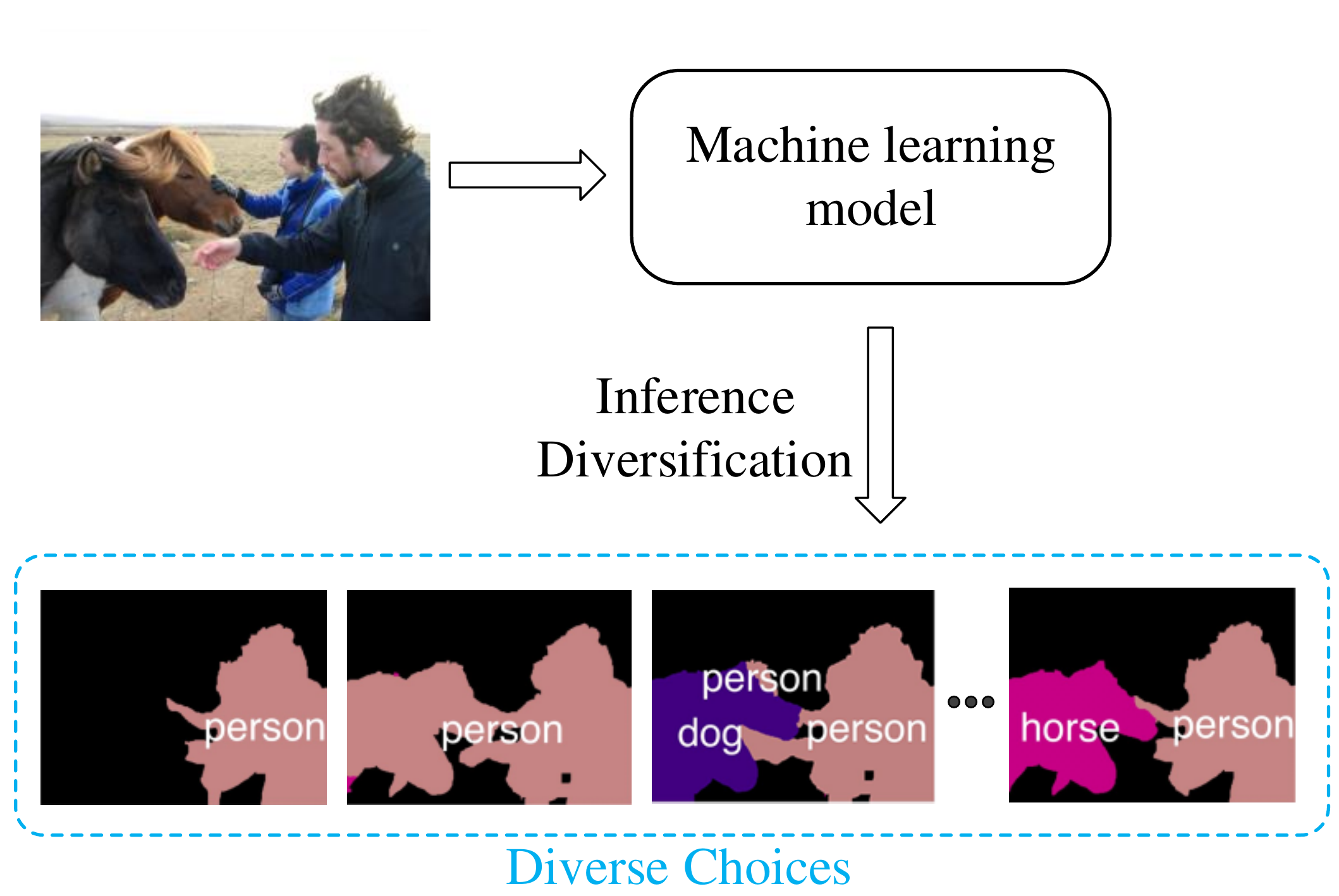}
\caption{Effects of inference diversification for improving the performance of the machine learning model.  The results come from prior work \cite{9}. Through inference diversification, multiple diversified choices can be obtained. Then, under the help of other methods, such as re-ranking in \cite{9}, the final solution can be obtained.}
\label{fig:inference_diversification}       
\end{figure}

We consider a set of discrete random variables  $X=\{x_i|i\in\{1,2,\cdots,N\}\}$, each taking value $y_i$ in a finite label set $L_v$. Let $G=(V,E)$($V=\{1,2,\cdots,N\}$, $E=\binom{V}{2}$) describe a graph defined over these variable. The set $L_\chi=\prod_{v\in \chi}L_v$ denotes a Cartesian product of sets of labels corresponding to the subset $\chi\in V$ of variables. Besides, denote $\theta_A:X_A\rightarrow R$, ($\forall A\in V\cup E$) as the  functions which define the energy at each node and edge for the labelling of variables in scope. The goal of the MAP inference is to find the labelling ${\bf y}=\{y_1,y_2,\cdots,y_N\}$ of the variables that minimizes this real-valued energy function:
\begin{equation}
{\bf y}^*=\arg\min\limits_{\bf y}E({\bf y})=\arg \min\limits_{{\bf y}}\sum_{A\in V\cup E} \theta_A({\bf y})
\end{equation}
However, ${\bf y}^*$ usually converges to the sub-optimal results due to the limited representational ability of the model and the limited training samples. Therefore, multiple choices, which can provide complement information, are desired from the model for the specialist.
Traditional methods to obtain multiple choices ${\bf y}^1, {\bf y}^2,\cdots,{\bf y}^M$ try to solve the following optimization:
\begin{equation} \label{eq:81}
\begin{aligned}
&{\bf y}^m=\arg\min\limits_{\bf y}E({\bf y})=\arg\min\limits_{\bf y}\sum_{A\in V\bigcup E}\theta_A({\bf y})\\
&s.t. \ {\bf y}^m\neq {\bf y}^i, i=1,2,\cdots,m-1
\end{aligned}
\end{equation}
However, the obtained second-best choice will typically be one-pixel shifted versions of the best \cite{23}. In other words, the next best choices will almost certainly be located on the upper slope of the peak corresponding with the most confident detection, while other peaks may be ignored entirely.

To overcome this problem, many methods, such as diversified multiple choice learning (D-MCL), submodular, M-Modes, M-NMS, have been developed for inference diversification in prior works. These methods try to diversify the obtained choices (do not overlap under a user-defined criteria) while obtaining high score on the optimization term.
Fig. \ref{fig:inference_diversification} shows some results of image segmentation from \cite{9}. Under the inference diversification, we can obtain multiple diversified choices, which represent the different optima of the data. Actually, {there also exist many methods which focus on providing multiple diversified choices in the inference phase.} In this work, we summarize the diversification in these works as inference diversification. The following subsections will introduce these works in detail.

\begin{table}
\centering
\caption{Overview of most frequently used inference diversification methods and the papers in which example measurements can be found.}
\label{table:03}       
\begin{tabular}{p{0.2\textwidth}p{0.22\textwidth}}
\hline\noalign{\smallskip}
Measurements & Papers   \\
\noalign{\smallskip}\hline\noalign{\smallskip}
D-MCL &  {\cite{4,5,7,9,11,16, 87}} \\
Submodular for Diversification & {\cite{6,8,94, add_34,add_35, add_44,add_45}}  \\
M-modes & \cite{12}  \\
M-NMS & {\cite{19,20,21,89, 96, 97}}  \\
DPP & {\cite{32, 152, add_8}} \\
\noalign{\smallskip}\hline
\end{tabular}
\end{table}

\subsection{Diversity-Promoting Multiple Choice Learning (D-MCL)}\label{subsec:mcl}
The D-MCL tries to find a diverse set of highly probable solutions under a discrete probabilistic model. Given a dissimilarity function measuring similarity between the pairwise choices, our formulation involves maximizing a linear combination of the probability and the dissimilarity to the previous choices. Even if the MAP solution alone is of poor quality, a diverse set of highly probable hypotheses might still enable accurate predictions. The goal of D-MCL is to produce a diverse set of low-energy solutions.

The first method is to approach the problem with a greedy algorithm, where the next choice is defined as the lowest energy state with at least some minimum dissimilarity from the previously chosen choices. To do so, a dissimilarity function $\Delta({\bf y},{\bf y}^i)$ is defined first. In order to find the $M$ diverse, low energy, labellings ${\bf y}^1, {\bf y}^2, \cdots, {\bf y}^M$, the method proceeds by solving a sequence of problems of the form \cite{5, 9, 11, 16, 87}
\begin{equation} \label{eq:65}
{\bf y}^m=\arg\min\limits_{\bf y}(E({\bf y})-\gamma\sum_{i=1}^{m-1}\Delta({\bf y},{\bf y}^i))
\end{equation}
for $m=1,2,\cdots,M$, where $\gamma>0$ determines a trade-off between diversity and energy, ${\bf y}^1$ is the MAP-solution and the function $\Delta:L_V\times L_V\rightarrow R$ defines the diversity of two labels. In other words, $\Delta({\bf y},{\bf y}^i)$ takes a large value if ${\bf y}$ and ${\bf y}^i$ are diverse, and a small value otherwise. For special case, the M-Best MAP is obtained when $\Delta$ is a 0-1 dissimilarity (i.e. $\Delta({\bf y}, {\bf y}^i)=I({\bf y}\neq {\bf y}^i)$).
The method considers the pairwise dissimilarity between the obtained choices. {More importantly, it is easy to understand and implement.} However, under the greedy strategy, each new labelling is obtained based on the previously found solutions, and ignores the upcoming labellings \cite{7}.

Contrary to the former form, the second method formulate the $M$-best diverse problem in form of a single energy minimization problem \cite{7}. Instead of the greedy sequential procedure in (\ref{eq:65}), this method suggests to infer all $M$ labellings jointly, by minimizing
\begin{equation} \label{eq:82}
E^M({\bf y}^1,{\bf y}^2,\cdots,{\bf y}^M)=\sum_{i=1}^{M}E({\bf y}^i)-\gamma \Delta^M({\bf y}^1,{\bf y}^2,\cdots,{\bf y}^M)
\end{equation}
where $\Delta^M$ defines the total diversity of any $M$ labellings. To achieve this, let us first create $M$ copies of the initial model. Three specific different diversity measures are introduced. The split-diversity measure is written as the sum of pairwise diversities, i.e. those penalizing pairs of labellings \cite{7}
\begin{equation}
\Delta^M({\bf y}^1,{\bf y}^2,\cdots,{\bf y}^M)=\sum_{i=2}^{M}\sum_{j=1}^{i-1}\Delta({\bf y}^i,{\bf y}^j)
\end{equation}
The node-diversity measure is defined as \cite{7}
\begin{equation}
\Delta^M({\bf y}^1,{\bf y}^2,\cdots,{\bf y}^M)=\sum_{v\in V}\Delta_v(y_v^1,y_v^2,\cdots,y_v^M)
\end{equation}
Finally, the special case of the split-diversity and node-diversity measures is the node-split-diversity measure \cite{7}
\begin{equation}
\Delta^M({\bf y}^1,{\bf y}^2,\cdots,{\bf y}^M)=\sum_{v\in V}\sum_{i=2}^{M}\sum_{j=1}^{i-1}\Delta_v(y_v^i,y_v^j)
\end{equation}

The D-MCL methods try to find multiple choices with a dissimilarity function. This can help the machine learning model to provide choices with more difference and show more diversity. However, the obtained choices may not be the local extrema  and {there may exist other choices} which could better represent the objects than the obtained ones.

\subsection{Submodular for Diversification }\label{subsec:submodular_mcl}
The problem of searching for a diverse but high-quality subset of items in a ground set $V$ of $N$ items has been studied in {information retrieval \cite{add_35}, web search \cite{add_34}, social networks \cite{add_47}, sensor placement \cite{add_48}, observation selection problem \cite{add_46}, set cover problem \cite{add_49}, document summarization \cite{add_44,add_45}, and others}. In many of these works, an effective, theoretically-grounded and practical tool for measuring the diversity of a set $S\subseteq V$ are submodular set functions. Submodularity is a property that comes from marginal gains. A set function $F:2^V\rightarrow R$ is submodular when its marginal gains $F(a|S)\equiv F(S\cup a)-F(S)$ are decreasing: $F(a|S)\geq F(a|T)$ for all $S\subseteq T$ and $a\notin T$. In addition, if  $F$ is monotone, i.e. $F(S)\leq F(T)$ whenever $S\subseteq T$, then a simple greedy algorithm that iteratively picks the element with the largest marginal gain $F(a|S)$ to add to the current set $S$, achieves the best possible approximation bound of $\displaystyle{(1-\frac{1}{e})}$\cite{94}. This result has presented significant practical impact. Unfortunately, if the number $N=|V|$ of items is exponentially large, then even a single linear scan for greedy augmentation is simply infeasible. The diversity is measured by a monotone, nondecreasing and normalized submodular function $f:2^V\rightarrow R^+$.

Denote $S$ as the set of choices. The diversification is measured by a monotone, nondecreasing and normalized submodular function $D: 2^V\rightarrow R^+$ . Then, the problem can be transformed to find a maximizing {configurations for the combined score \cite{6, add_34, add_35, add_44, add_45}}
\begin{equation}
F(S)=E(S)+\gamma D(S)
\end{equation}
The optimization can be solved by the greedy algorithm that starts out with $S^0=\emptyset$, and {iteratively adds the best term:}
\begin{equation}
\begin{aligned}
{\bf y}^m=&\arg\max\limits_{\bf y}F({\bf y}|S^{m-1}) \\
=&\arg\max\limits_{\bf y}\{E({\bf y})+\gamma D({\bf y}|S^{m-1})\}
\end{aligned}
\end{equation}
where $S^m={\bf y}^m\bigcup S^{m-1}$. The selected choice $S^m$ is within a factor of $\displaystyle{1-\frac{1}{e}}$ of the optimal solution $S^*$:
\begin{equation}
F(S^m)\geq (1-\frac{1}{e})F(S^*).
\end{equation}

The submodular takes advantage of the maximization of marginal gains to find multiple choices which can provide the maximum of complement information.

\subsection{M-NMS}\label{subsec:nms}

Another way to obtain multiple diversified choices is the non-maximum suppression (M-NMS) \cite{150, 97}. The M-NMS is typically defined in an algorithmic way: starting from the MAP prediction one goes through all labellings according to an increasing order of the energy. A labelling becomes part of the predicted set if and only if it is more than $\rho$ away from the ones chosen before, where $\rho$ is the threshold defined by user to judge whether two labellings are similar. The M-NMS guarantee the choices to be apart from each other. The M-NMS is typically implemented by greedy algorithm \cite{20, 89, 96, 97}.

A simple greedy algorithm for instantiating multiple choices are used: Search over the exponentially large space of choices for the maximally scoring choice, instantiate it, {remove all choices with overlapping, and repeat}. The process is repeated until the score for the next-best choice is below a threshold or M choices have been instantiated. However, {the general implementation} of such an algorithm would take exponential time.

The M-NMS method tries to find M-best choices by throwing away the similar choices from the candidate set. To be concluded, the D-MCL, submodular, and M-NMS have the similar idea. All of them tries to find the M-best choices under a dissimilarity function or the ones which can provide the most complement information.
\subsection{M-modes}\label{subsec:modes}

Even though the former three methods guarantee the obtained multiple choices to be apart from each other, the choices are typically not local extrema of the probability distribution. {To further guarantee both the local extrema and the diversification of the obtained multiple choices simultaneously, the problem can be transformed to the M-modes.} The M-modes have multiple possible applications, because they are intrinsically diverse.

For a non-negative integer $\delta$, define the  $\delta$-neighborhood of a labelling ${\bf y}$ to be $N_\delta=\{{\bf y}|d({\bf y},{\bf y}')\leq \delta\}$ as the set of labellings whose distances from ${\bf y}$ is no more than $\delta$, where $d(\cdot)$ measures the distance between two labellings, such as the Hamming distance.
Then, a labelling ${\bf y}$ is defined as a local maximum of {the energy function} $E(\cdot)$, iff $ E({\bf y})\geq E({\bf y}')$, $\forall {\bf y}'\in N({\bf y})$.


Given $\delta$,  the set of modes is denoted by $M^\delta$, formally, \cite{12}
\begin{equation}
  M^\delta=\{{\bf y}| E({\bf y}')\geq E({\bf y}), \forall {\bf y}'\in N_\delta({\bf y})\}
\end{equation}
As $\delta$ increases from zero to infinity, the $\delta$-neighborhood of ${\bf y}$ monotonically grows and the set of modes $M^\delta$ monotonically decreases. Therefore, the $M^\delta$ can form a nested sequence, \cite{12}
\begin{equation}
M^0\supseteq M^1\supseteq \cdots \supseteq M^\infty=\{{\text{MAP}}\}
\end{equation}
{Here, the M-modes} can be defined as computing the $M$ labellings with minimal energies in $M^\delta$.
{Then, the problem has been transformed to M-modes:} Compute the $M$ labellings with minimal energies in $M^\delta$.

Besides, \cite{12} has already validated that a labelling is a mode if and only if it behaves like a "local mode" everywhere, and thus a new chain has been constructed and M-modes problem is reduced into the $M$ best problem of the new chain.

Furthermore, it also validates the one-to-one cost preserving correspondence between consistent configurations $\alpha$ and the set of modes $M^\delta$. Therefore, the problem of computing the $M$ best modes are transferred to the problem of computing the $M$ best configurations in the new chain.

Different from the former three methods, M-modes can obtain M choices which are the local extrema of the optimization and this can provide M choices which contains the most complement information.

\subsection{DPP}\label{subsec:dpp_mcl}

General M-NMS and M-Modes try to select choices with the highest scores. Since these methods give a priority to scores of the choices, it might finally end up in selecting the  overlapping choices and miss the best possible set of non-overlapping ones with acceptable scores {\cite{32, 152, add_8}}. To address this problem, {\cite{32, 152, add_8}} attempt to use the DPP to select a set of diverse and informative choices with enriched representations.


The definition of DPP have been introduced in detail in subsection \ref{subsec:dpp}.
It can be noted that the DPP is a distribution over the subsets of a fixed ground set, which prefers a diverse set of points.
The selected subset of items by DPP can be representative and cover significant amount of information from the whole set. Besides, the selection would be diverse and non-repetitive {\cite{32, add_8}}.
To make the inference diversification, \cite{32} calculates the probability of inclusion of each choice depends on the determinant of a kernel matrix. The kernel matrix is defined such that it captures all spatial and contextual information between choices all at once. To apply the DPP, the quality and diversity terms need to be defined. The quality term (unary score) defines the optimization term, such as the $E(y)$ in Eq. \ref{eq:81}. The diversity term defines the pairwise correlation of the obtained choices.

Similar to Eq. \ref{eq:82}, the model is first transformed into the M-best diverse problem in form of a single energy minimization problem. The optimization problem based on the DPP can be formulated as
\begin{equation} \label{eq:83}
E^M({\bf y}^1,{\bf y}^2,\cdots,{\bf y}^M)=\sum_{i=1}^{M}E({\bf y}^i)-\gamma \det(L({\bf y}^1,{\bf y}^2,\cdots,{\bf y}^M))
\end{equation}
The kernel matrix in DPP is defined as
\begin{equation}
L({\bf y}^1,{\bf y}^2,\cdots,{\bf y}^M)=[L_{ij}]_{i,j=1,2,\cdots,M}
\end{equation}
where $L_{ij}=E({\bf y}^i)E({\bf y}^j)S_{ij}$. $S_{ij}$ is the similarity term to computer the dissimilarity between different choices.
A DPP tends to pick uncorrelated choices in these cases. On the other hand, higher quality items increase the determinant, and thus a DPP tends to pick high-quality and diverse set of choices.

\subsection{Analysis}
Even though all the methods in former subsections can be used for inference diversity, there exists some difference between these methods. These methods in prior works are summarized in Table \ref{table:03}. It can be noted from the former subsections that the D-MCL is {the easiest one to implement}. One only needs to calculate the MAP choice and obtain other choices by constraining the optimization with a dissimilarity function. In contrary, the M-NMS neglects the choices which are in the neighbors of the former choice and obtain other choices from the remainder. The D-MCL and M-NMS obtain choices by solving the optimization with the user-defined similarity while the submodular method tries to obtain the choices which can provide the maximal marginal and complement information. The former three methods may provide choices which is not local optima while  the local optimal choices usually contain more information than others. Therefore, different from the former three methods, the M-modes tries to obtain multiple diversified choices which are also local optima.
All these methods before can be used in traditional machine learning methods. The DPP method for inference diversification which is developed by \cite{32, 152} is mainly applied in object detection tasks. The methods in \cite{32, 152} take advantage of the merits of DPP to obtain high quality choices with less overlapping which could present better performance than M-NMS and M-modes. From the introduction of data diversification, D-model, D-models, and inference diversification, one can choose the proper method for diversification of machine learning in various computer vision tasks. In the following, we'll introduce some applications of diversity technology in machine learning model.

\section{Applications}\label{sec:application}

Diversity technology in machine learning can significantly improve the representational ability of the model in many computer vision tasks, including {the remote sensing imaging tasks \cite{13, 14, 74, 123}, camera relocalization \cite{15, 104}, natural image segmentation \cite{9, 5, 11}, object detection \cite{19, 20}, machine translation \cite{16, 124}, information retrieval \cite{33, add_41, add_42, add_43, add_35}, social network analysis \cite{add_36, add_35, add_39}, document summarization \cite{add_44, add_45, add_51}, web search \cite{add_34,add_40, add_38, add_53}, and others}. The diversity priors, which can decrease the redundancy in the learned model or diversify the obtained multiple choices, can provide more informative features and show powerful ability in real-world application, especially for the {machine learning tasks} with limited training samples and complex structures in the training samples. In the following, we'll introduce some of the applications of the diversity technology in machine learning.

\subsection{Remote Sensing Imaging Tasks}

Remote sensing images, such as the hyperspectral images and the multi-spectral images, have played a more and more important role in the past two decades \cite{101}. However, there exist some typical difficulties in remote sensing imaging tasks. First, limited number of training samples in remote sensing imaging tasks usually make it difficult to {describe the images}. Since labelling is usually time-consuming and cost, it usually cannot provide enough training samples to train the model. Besides, remote sensing images usually have large intra-class variance and low inter-class variance, which make it difficult to extract discriminative features from the images.
Therefore, proper feature extraction models are required for the representations of the remote sensing images. Recently, deep models have demonstrated their impressive performance in extracting features from the remote sensing images \cite{153}. However, the deep models usually consist of large amounts of parameters while the limited training samples would make the learned deep model be sub-optimal. This would limit the performance of the deep models for the representation of the remote sensing images.

To overcome these problems, some works have applied the diversity-promoting prior to diversify the model for better performance \cite{13, 14, gong_cnn, 154}. \cite{13} and \cite{14} attempt to diversify the learned model with the independence prior, which is based on the cosine similarity in the former section for remote sensing images.  \cite{13} develops a special diversity-promoting deep structural metric learning method for scene classification in remote sensing  while \cite{14} imposes the independence prior  on deep belief network (DBN) for hyperspectral image classification. If we denote $W=[{\bf w}_1, {\bf w}_2, \cdots, {\bf w}_s]$ as the metric parameter factors in \cite{13} and the latent factors of RBM in \cite{14}, then the diversity term by the independence prior in the two papers can be formulated as
\begin{equation}
f(W)=\sum^{K}_{i\neq j}\frac{<{\bf w}_i,{\bf w}_j>}{\|{\bf w}_i\|\|{\bf w}_j\|}
\end{equation}
{where $K$ is the number of the factors.}
The diversity term $f(W)$ encourages the factors to be diversified, so as to improve the representational ability of the model for the images. As introduced in subsection \ref{subsec:diversity_regularization}, to make use of the multiple information, \cite{gong_cnn} have applied the DPP prior in the learning process of deep model for hyperspectral image classification. The DPP prior can be formulated as
\begin{equation}
P(W)=(\det(\psi(W)))^\gamma
\end{equation}
The diversity regularization $f(W)$ in Eq. \ref{eq:47} can be written as
\begin{equation}
f(W)=-\log[(\det(\psi(W)))]
\end{equation}
{Besides, \cite{gong_cnn} also provides the way to process the diversity regularization by the back propagation,}
\begin{equation}
\frac{\partial f(W)}{\partial W}= - \frac{\partial \log \det(\psi(W))}{\partial W}=-2 W(WW^T)^{-1}.
\end{equation}
With the developed diversified model for hyperspectral image representation, the classification performance can be significantly improved \cite{gong_cnn}.
These prior works mainly improve the representational ability of the model for better performance from the D-model view.

Besides, \cite{154, gong_ijcnn} tries to improve the representational ability from the data diversification way. \cite{154, gong_ijcnn} create the pseudo classes with the center points. In \cite{154}, the pseudo classes are used to decrease the intra-class variance. Furthermore, the diversity-promoting prior, which is created based on the Euclidean distance to repulse different pseudo classes from each other, is used to improve the effectiveness of the developed training process for remote sensing scenes.
In \cite{gong_ijcnn}, the pseudo classes are used for unsupervised learning of remote sensing scene representation. The pseudo classes are used to allocate pseudo labels and training the model under the supervised way. Similar to \cite{154}, the diversity-promoting prior is also used to repulse different pseudo classes from each other to improve the effectiveness of the unsupervised learning process.

Furthermore, some other works focus on the diversification of multiple models for remote sensing images \cite{74, 123}.  \cite{74} has applied cross entropy measurement to diversify the obtained multiple models and then the obtained multiple models could provide more complement information (See subsection \ref{subsubsec:optimization} for details). Different from \cite{74},  \cite{123} divides the training samples  into several subsets for different models, separately. Then, each model focuses on the representation of different classes and the whole representation of these models could be improved (See subsection \ref{subsubsec:sample} for details).

\begin{table*}
\centering
\caption{{Some comparison results between the general model and the diversified model for remote sensing imaging tasks.}}
\label{table:comparison_result}       
\begin{tabular}{c | c | c c | c c}
\hline\noalign{\smallskip}
Dataset & Reference & Methods & Accuracy (\%) & Diversified Method & Accuracy (\%)  \\
\noalign{\smallskip}\hline\noalign{\smallskip}
\multirow{2}*{{Ucmerced Land Use dataset}} & \cite{13} & DSML & $95.95 \pm 0.24$ & D-DSML & $96.76 \pm 0.36$ \\
& \cite{74} & CaffeNet & $95.48$ & Diversified MCL & $97.05 \pm 0.55$  \\
\hline
\multirow{3}*{{Pavia University}} & \cite{14} & DBN & $91.18 \pm 0.08$ & D-DBN-PF & $93.11 \pm 0.06$ \\
& \cite{gong_cnn} & DML-MS-CNN & $99.03 \pm 0.25$ & DPP-DML-MS-CNN & $99.46 \pm 0.03$  \\
& \cite{123} & DBN & $90.61 \pm 1.15$ & M-DBN & $92.55 \pm 0.74$ \\
\hline
\multirow{2}*{{Indian Pines}} & \cite{14} & DBN & $88.25 \pm 0.17$ & D-DBN & $91.03 \pm 0.12$ \\
& \cite{gong_cnn} & DML-MS-CNN & $98.87 \pm 0.21$ & DPP-DML-MS-CNN & $99.08 \pm 0.23$  \\
\noalign{\smallskip}\hline
\end{tabular}
\end{table*}

{To further describe the effectiveness of the diversity-promoting methods in machine learning, we listed some comparison results between the general model and the diversified model over different datasets in Table \ref{table:comparison_result}. The results come from the prior works. This work choose the Ucmerced Land use dataset, Pavia University dataset, and the Indian Pines as representatives.}

{Ucmerced Land Use dataset \cite{dataset_1} was manually extracted from orthoimagery. It is multi-class land-use scenes in the visible spectrum which contains 2100 aerial scene images divided into 21 challenging scene categories, including agricultural, airplane, baseball diamond, beach, buildings, chaparral, dense residential, forest, freeway, golf course, harbor, intersection, medium density residential, mobile home park, overpass, parking lot, river, runway, sparse residential, storage tanks, and tennis court. Each scene has $256 \times 256$ pixels with a resolution of one foot per pixel. For the experiments in Table \ref{table:comparison_result}, 80\% scenes of each class are used for training and the remainder are for testing.}

{Pavia University dataset \cite{dataset_2} was gathered by a sensor known as the reflective optics system imaging spectrometer (ROSIS-3) over the city of Pavia, Italy. The image consists of $610 \times 340$ pixels with 115 spectral bands. The image is divided into 9 classes with a total of 42, 776 labelled samples, including the asphalt, meadows, gravel, trees, metal sheet, bare soil, bitumen, brick, and shadow. For the experiments, 200 samples of each class are used for training and the remainder are for testing.}

{Indian Pines dataset \cite{dataset_3} was taken by AVIRIS sensor in northwestern Indiana. The image has $145 \times 145$ pixels with 224 spectral channels where 24 channels are removed due to the noise. The image is divided into 8 classes with a total of 8, 598 labelled samples, including the Corn no\_till, Corn min\_till, Grass pasture, hay windrowed, Soybeans no\_till, Soy beans min, Soybeans clean, and woods. For the experiments, 200 samples of each class are used for training and the remainder are for testing.}

{From the comparisons in Table \ref{table:comparison_result}, we can find that the diversity technology can improve the representational ability of the machine learning model and thus significantly improve the classification performance of the learned machine learning model.}

\subsection{Image Segmentation}

In computer vision, image segmentation is the process of partitioning a digital image into multiple segments (sets of pixels). The goal of segmentation is to simplify and change the representation of an image into something that is more meaningful and easier to analyze. More precisely, image segmentation is the process of assigning a label to each pixel in an image such that pixels with the same label share certain characteristics. Since a semantic segmentation algorithm deals with tremendous amount of uncertainty from inter and intra object occlusion and varying appearance, lighting and pose, obtaining multiple best choices from all possible segmentations tends to be one of the possible way to solve the problem. Therefore, the image segmentation problem can be transformed into the M-best problem. However, as traditional problem in M-best problem, the obtained multiple choices are usually similar and the information provided to the user tends to be redundant.
As section \ref{sec:inference} shows, the obtained multiple choices will usually be  only one-pixel shifted versions to each other.

The way to solve this problem is to introduce diversity into the training process to encourage the multiple choices to be diverse. {Under inference diversification,} the model is desired to provide a diverse set of low-energy solutions which represent different local optimal results from the  data. Fig. \ref{fig:inference_diversification} shows the examples of inference diversification over the task in prior work \cite{9}.
Many works \cite{9, 5, 11, 4, 7, 6, 125, 126, 127} have introduced inference diversity into the image segmentation tasks via different ways. \cite{5} first introduces the D-MCL in subsection \ref{subsec:mcl} for image segmentation. The work developed the diversification method as Eq. \ref{eq:65} shows. To further improve the performance in work \cite{5},
prior works \cite{9} and \cite{4} combine the D-MCL with reranking which provide a way to obtain multiple diversified choices and select the proper one from the multiple choices.
As discussed in subsection \ref{subsec:mcl}, the greedy nature of original D-MCL makes the obtained labelling only be influenced by previously found labellings while ignore the upcoming labellings. To overcome this problem,  \cite{7} develops a novel D-MCL which has the form as Eq. \ref{eq:82}.
Besides, \cite{6, 127} use the submodular to measure the diversification between multiple choices (see details in subsection \ref{subsec:submodular_mcl}).
\cite{126} combines the NMS (see details in subsection \ref{subsec:nms}) and the sliding window to obtain multiple choices.
Instead of inference diversification for better performance of image segmentation, prior work \cite{99} tries to obtain multiple diversified models for image segmentation task.
The method proposed by \cite{99} is to divide the training samples into several subsets where each base model is trained with a specific one. Through allocating each training sample to the model with lowest predict error, each model tends to model different classes from others.
Under the inference diversification and D-models for image segmentation tasks, the obtained multiple choices would be diversified as Fig. \ref{fig:inference_diversification} shows and the performance of the model would also be significantly improved.

\begin{table*}
\centering
\caption{{Some comparison results between the general model and the diversified model for image segmentation.}}
\label{table:comparison_result_segmentation}       
\begin{tabular}{c | c | c c | c c}
\hline\noalign{\smallskip}
Dataset & Reference & Methods & Accuracy (\%) & Diversified Method & Accuracy (\%)  \\
\noalign{\smallskip}\hline\noalign{\smallskip}
{{PASCAL VOC 2010 dataset}} & \cite{5} & MAP & $91.54$ & DivMBEST & $95.16$ \\
{{PASCAL VOC 2011 dataset}} & \cite{99} & MCL & about 66 & sMCL & about 71  \\
{{PASCAL VOC 2012 dataset}} & \cite{9} & Second Order Pooling ($O_2P$)-MAP & $46.5$ & DivMBEST+Ranking & $48.1$ \\
{{PASCAL VOC 2012 dataset}} & \cite{6} & MAP & $43.43$ & submodular-MCL & $55.32$  \\
\noalign{\smallskip}\hline
\end{tabular}
\end{table*}

{Just as the remote sensing imaging tasks, we list some comparison results in prior works to show the effectiveness of the diversity technology in machine learning. The comparison results are listed in Table \ref{table:comparison_result_segmentation}. Generally, the experimental results on the PASCAL Visual Object Classes Challenge (PASCAL VOC) dataset are chosen to show the effectiveness of the diversity in machine learning for image segmentation.}

{Also, From the table \ref{table:comparison_result_segmentation}, we can find that the inference diversification can significantly improve the performance for segmentation.}

\subsection{Camera Relocalization}

Camera relocalization is to estimate the pose of a camera relative to a known 3D scene from a single RGB-D frame \cite{104}. It can be formulated as the inversion of the generative rendering procedure, which is to find the camera pose corresponding to a rendering of the 3D scene model that is most similar to the observed input. Since the problem is a non-convex optimization problem which has many local optima, one of the methods to solve the problem is to find a set of M predictors which generate M camera pose hypotheses and then infers the best pose from the multiple pose hypotheses. Similar to traditional M-best problems, the obtained M predictors is usually similar.

To overcome this problem and obtain hypotheses that are different from each other, \cite{15} tries to learn 'marginally relevant' predictors, which can make complementary predictions, and compare their performance when used with different selection procedures. In \cite{15}, greedy algorithm is used to obtain multiple diversified models. Different weights are defined on each training samples, and the weights is updated with the training loss from the former learned model. Finally, multiple diversified models can be obtained for camera relocalization.

\subsection{Object Detection}

Object detection is one of the computer vision tasks which deals with detecting instances of semantic objects of a certain class (such as humans, buildings, or cars) in digital images and videos. Similar to image segmentation tasks, great uncertainty is contained in the object detection algorithms. Therefore, obtaining multiple diversified choices is also an important way to solve the problem.

Some prior works \cite{19, 20} have made great effects to obtain multiple diversified choices by M-NMS (see \ref{subsec:nms} for details). \cite{20} use the greedy procedure for eliminating repeated detections via NMS. Besides, \cite{19} demonstrates that the energies resulting from M-NMS lead to the maximization of submodular function, and then through branch-and-bound strategy \cite{100}, all the image can be explored and diversified multiple detections can be obtained.

\subsection{{Machine Translation}}

Machine translation (MT) task is a sub-field of computational linguistics that investigates the use of software to translate text or speech from one language to another. Recently, machine translation systems have been developed and widely used in real-world application. Commercial machine translation services, such as Google translator, Microsoft translator, and Baidu translator, have made great success. From the perspective of the user interaction, the ideal machine translator is an agent that reads documents in one language and provides accurate, high quality translations in another. This interaction ideal has been implicit in machine translation (MT) research since the field's inception. Unfortunately, when a real, imperfect MT system makes an error, the user is left trying to guess what the original sentence means. Therefore, to overcome this problem, providing the M-best translations instead of a single best one is necessary \cite{102}.

However, in MT, for example, many translations on M-best lists are extremely similar, often differing only by a single punctuation mark or minor morphological variation. The implicit goal behind these technologies is to better explore the output space by introducing diversity into the surrogate set.
Some prior works have introduced diversity into the obtained multiple choices and obtained better performance \cite{124}.  \cite{16} develops the method to diversify multiple choices which is introduced in subsection \ref{subsec:mcl}. In the works, a novel dissimilarity function has been defined on different translations to increase the diversity between the obtained translations. It can be formulated as \cite{16}
\begin{equation}
  \Delta_n(y,y')=-\sum_{i=1}^{|y|-q}\sum_{j=1}^{|y'|-q}[[y_{i:i+q}=y'_{j:j+q}]]
\end{equation}
where $[[\cdot]]$ is the Iverson bracket (1 if input condition is true, 0 otherwise) and $y_{i:j}$ is the subsequence of $y$ from word $i$ to word $j$ (inclusive). The advantage of this dissimilarity function is its simplicity. Besides, the diversity-promoting can ensure the machine learning system obtain multiple diversified translation for the user.

\subsection{{Information Retrieval}}

\subsubsection{{Natural Language Processing}}
In machine learning and natural language processing, a topic model is a statistical model for discovering the abstract "topics" that occur in a collection of documents. Probabilistic topic models such as Latent Dirichlet Allocation (LDA) and Restricted Boltzmann Machine (RBM) can provide a useful and elegant tool for discovering hidden structure within large data sets of discrete data, such as corpuses of text. However, LDA implicitly discovers topics along only a single dimension while RBM tends to learn multiple redundant hidden units to best represent dominant topics and ignore those in the long-tail region \cite{26}. To overcome this problem, diversifying over the learned model (D-model) can be applied over the learning process of the model.

Recent research on multi-dimensional topic modeling aims to devise techniques that can discover multiple groups of topics, where each group models some different dimension or aspect of the data.
Therefore, prior work \cite{33} presents a new multi-dimensional topic model that uses a determinantal point process prior (see details in subsection \ref{subsec:mcl}) to encourage different groups of topics to model different dimensions of the data. Determinantal point processes are probabilistic models of repulsive phenomena which originated in statistical physics but have recently seen interest from the machine learning community.
Besides, \cite{26} introduces the RBM for topic modeling to utilize hidden units to discover the latent topics and learn compact semantic representations for documents. Furthermore, to reduce the redundancy  of the learned RBM, \cite{26} utilizes the angular-based diversification method which has the form of Eq. \ref{eq:84} to diversify the learned hidden units in RBM. Under the diversification, the RBM can learn much more powerful latent document representations that boost the performance of topic modeling greatly \cite{26}.

\subsubsection{{Web Search}}

{The problem of result diversification has been studied in various tasks, but the most robust literature on result diversification exists in web search \cite{add_53}. Web search has become the prodominant method for people to fulfill their information needs. In web search, it is general to provide different search results with different interpretations of a query. To satisfy the requirement of multiple distinct user type, the web search system is desired to provide a diverse set of results.

The increasing requirements for easily accessible information via web-based services have attracted a lot of attentions on the studies of obtaining diverse search results \cite{add_34,add_40, add_38, add_53}. The objective is to achieve large coverage on a few features but very small coverage on the remaining features, which satisfies the submodularity. Therefore, \cite{add_34,add_53} take advantage of the submodular for result diversification (see Subsection \ref{subsec:submodular_mcl} for details). Besides, \cite{add_38} uses the distance-based measurement to formulate the dissimilarity function in Subsection \ref{subsec:mcl} and \cite{add_40} has further summarized the diversification methods for result diversification which has the form as D-MCL (See Subsection \ref{subsec:mcl} for details). These search diversification methods can provide multiple diversified choices for the users to satisfy the requirements of specific information.}

\subsection{{Social Network Analysis}}

{Social network analysis is a process of investigating social structures through the use of networks and graph theory. It characterizes networked structures in terms of nodes and the ties, edges, or links (relationships or interactions) that connect them. Ranking nodes on graphs is a fundamental task in social network analysis and it can be applied to measure the centrality in social networks \cite{add_35}. However, many nodes in top-K ranking list obtained general methods are general similar since it only takes the relevance of the nodes into consideration \cite{add_36}.

To improve the effectiveness of the ranking process, many prior works have incorporated the diversity into the top-K ranking results \cite{ add_36, add_35, add_39}. To enforce diversity in the top-K ranking results, the way to measure the similarity tends to be the key problem. \cite{add_35} has formulated the diversified ranking problem as a submodular set function maximization and processes the problem as Subsection \ref{subsec:submodular_mcl} shows. Besides, \cite{add_36} takes advantage of the Heat kernel to formulate the weights on the social graph (see Subsection \ref{subsec:diversity_regularization} for details)
where larger weights would be if points are closer. Then through the random walks in an absorbing Markov chain, diversified ranking can be obtained \cite{add_36}. Furthermore, according to the special characteristics, \cite{add_39} develops a novel goodness measure to balance both the relevant and the diversity. For simplicity, the goodness measure would not be shown in this work and more details can be found in \cite{add_39} for interested readers. It should be noted that the optimization problem in \cite{add_39} is equal to the D-MCL in subsection \ref{subsec:mcl}.}

\subsection{{Document Summarization}}

{Multi-document summarization is an automatic procedure aimed at extraction of information from multiple texts written about the same topic. Generally, a good summary should coverage elements from distinct parts of data to be representative of the corpus and does not contain elements that are too similar to each other. Therefore, coverage and diversity are usually essential for the multi-document summarization \cite{add_50}. }

{Since coverage and diversity could sometimes be conflicting requirements \cite{add_50}, some prior works try to find a tradeoff between the coverage and the diversity. As Subsection \ref{subsec:submodular_mcl} shows, since there exist efficient algorithms (with near-optimal solutions) for a diverse set of constraints when the submodular function is monotone in the document summarization task, the submodular optimization methods have been applied in the diversification of the obtained summary of the multiple documents \cite{add_44, add_45, add_50, add_51, add_52}. \cite{add_44} defines a class of submodular functions meant for document summarization and \cite{add_45} treats the document summarization problem as maximizing a submodular function with a budget constraint. \cite{add_50} further investigates the personalized data summarization by the submodular function subject to multiple constraints. Under the diversification by submodular, the obtained elements can be diversified and the multiple documents can be better summarized.}

\section{Discussions}

This article surveyed the available works on diversity technology in general  machine learning model, by systematically categorizing the diversity in training samples, D-model, D-models, and inference diversity in the model. We first summarize the main results and identify the challenges encountered throughout the article.

Recently, due to the excellent performance of the machine learning model for feature extraction, machine learning methods have been widely applied in real-world applications. However, the limited number and imbalance of training samples in real-world applications usually make the learned machine learning models be sub-optimal, sometimes even lead to the "over-fitting" in the training process. This would limit the performance of the machine learning models. Therefore, this work summarizes the diversity technology in prior works which can work on the machine learning model as one of the methods to improve the model's representational ability. We want to emphasize that the diversity technology is not decisive. The diversity can only be considered as an additional technology to improve the performance of the machine learning process.

{Through the introduction of the diversity in machine learning, the three questions proposed in the introduction can be easily answered. The detailed descriptions of data diversification, model diversification, and inference diversification  are introduced in sections \ref{sec:data}, \ref{sec:model}, and \ref{sec:inference}. With these methods, the machine learning model can be diversified and the performance can be improved. Besides, the diversification of the model (D-model) tries to improve the representational ability of the machine learning model directly (see Fig. \ref{fig:d_model} for details) while the diversification of the models (D-models) aims to obtain multiple diversified choices under the diversification of the ensemble learning (see Fig. \ref{fig:d_models} for details). It should also be noted that the diversification measurements for data diversification, model diversification, and the inference diversification show some similarity. As introduced, the diversity aims to decrease the redundancy between the data or the factors.
The key problem for diversification is the way to measure the similarity between the data or the factors.
However, in the machine learning process, the data and the factors are processes as the vectors. Therefore, there exist overlaps between the data diversification, model diversification as well as the inference diversification, such as the DPP measurement. More importantly, we should also note that the diversification in different steps of machine learning models  presents its special characteristics. The details for the applications of diversity methods are shown in section \ref{sec:application}. This work only lists the most common applications in real-world. The readers should consider whether the diversity methods are necessary according to the specific task they face with.}

{\bf Advice for implementation.} We expect this article is useful to researchers who want to improve the representational ability of machine learning models for computer vision tasks. For a given computer vision task, the proper machine learning model should be chosen first. Then, we advise to consider adding diversity-promoting priors to improve the performance of the model and further what type of diversity measurement is desired. When one desires obtain multiple models or multiple choices, then one can consider diversifying multiple models or the obtained multiple choices and section \ref{subsec:d_models} and \ref{sec:inference} would be relevant and helpful. We advise the reader to first consider whether the multiple models or multiple choices can be helpful for the performance.

\section{Conclusions}

The training of machine learning models requires large amounts of labelled samples. However, the limited training samples constrain the performance of machine learning models. Therefore, effective diversity technology, which can encourage the model to be diversified and improve the representational ability of the model, is expected to be an active area of research in machine learning tasks. This paper summarizes the diversity technology for machine learning in previous works. We introduce diversity technology in data pre-processing, model training, inference, respectively.
Other researchers can judge whether diversity technology is needed and choose the proper diversity method for the special requirements according to the introductions in former sections.

\begin{IEEEbiography}[{\includegraphics[width=1in,height=1.25in,clip,keepaspectratio]{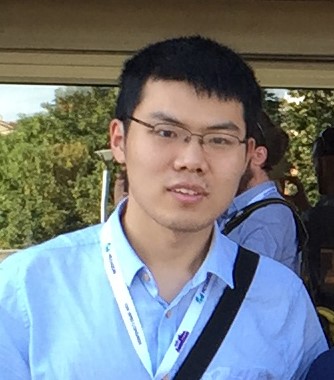}}]{Zhiqiang Gong}  received the Bachelor's degree in applied mathematics from Shanghai Jiaotong University (SJTU), Shanghai, China, in 2013, and the Mater degree in applied mathematics from National University of Defense Technology (NUDT), Changsha, China, in 2015. He is currently pursuing the Ph.D degree at National Key Laboratory of Science and Technology on ATR, National University of Defense Technology, Changsha, China. His research interests are computer vision and image analysis.
\end{IEEEbiography}

\begin{IEEEbiography}[{\includegraphics[width=1in,height=1.25in,clip,keepaspectratio]{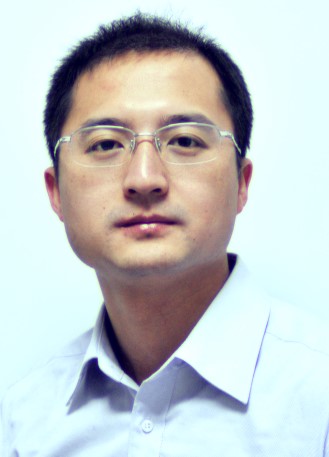}}]{Ping Zhong} (M'09--SM'18) received the M.S. degree in applied mathematics and the Ph.D. degree in information and communication engineering from the National University of Defense Technology (NUDT), Changsha, China, in 2003 and 2008, respectively.

He is currently an Associate Professor with the National Key Laboratory of Science and Technology on ATR, NUDT. From March 2015 to February 2016, he was a Visiting Scholar in the Department of Applied Mathematics and Theory Physics, University of Cambridge, Cambridge, U.K. He has authored or coauthored more than 30 peer reviewed papers in international journals such as the IEEE TRANSACTIONS ON NEURAL NETWORKS AND LEARNING SYSTEMS, the IEEE TRANSACTIONS ON IMAGE PROCESSING, the IEEE TRANSACTIONS ON GEOSCIENCE AND REMOTE SENSING, the IEEE JOURNAL OF SELECTED TOPICS IN SIGNAL PROCESSING, and the IEEE JOURNAL OF SELECTED TOPICS IN APPLIED EARTH OBSERVATIONS AND REMOTE SENSING. His current research interests include computer vision, machine learning, and pattern recognition.

Dr. Zhong is a Referee of the IEEE TRANSACTIONS ON NEURAL NETWORKS AND LEARNING SYSTEMS, the IEEE TRANSACTIONS ON IMAGE PROCESSING, the IEEE TRANSACTIONS ON GEOSCIENCE AND REMOTE SENSING, the IEEE JOURNAL OF SELECTED TOPICS IN SIGNAL PROCESSING, and the IEEE JOURNAL OF SELECTED TOPICS IN APPLIED EARTH OBSERVATIONS AND REMOTE SENSING, and the IEEE GEOSCIENCE AND REMOTE SENSING LETTERS. He was the recipient of the National Excellent Doctoral Dissertation Award of China (2011) and the New Century Excellent Talents in University of China (2013).

\end{IEEEbiography}

\begin{IEEEbiography}[{\includegraphics[width=1in,height=1.25in,clip,keepaspectratio]{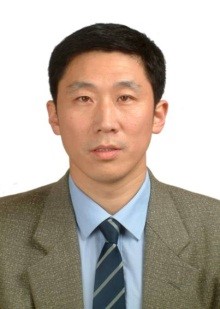}}]{Weidong Hu} was born in September, 1967. He received the B.S. degree in microwave technology, the M.S. degree and Ph.D. degree in communication and electronic system from the National University of Defense Technology, Changsha, P. R. China in 1990, 1994 and 1997, respectively.

He is currently a full professor with the National Key Laboratory of Science and Technology on ATR, National University of Defense Technology. His research interests include radar signal and data processing.

\end{IEEEbiography}

\EOD

\end{document}